\documentclass[10pt]{article}
\usepackage{times,latexsym}
\usepackage{url}
\usepackage[T1]{fontenc}
\usepackage{amssymb}
\usepackage{pifont}
\usepackage{graphicx}
\usepackage{adjustbox}
\usepackage{appendix}
\usepackage{hyperref}

\usepackage[accepted]{tmlr}

\title{A Survey on Compositional Learning of AI Models: Theoretical and Experimental Practices}
\author{\name Sania Sinha \email sinhasa3@msu.edu \\
      \addr Department of Computer Science and Engineering\\
      Michigan State University
      \AND
      \name Tanawan Premsri \email premsrit@msu.edu \\
      \addr Department of Computer Science and Engineering\\
      Michigan State University
      \AND
      \name Parisa Kordjamshidi \email kordjams@msu.edu\\
      \addr Department of Computer Science and Engineering\\
      Michigan State University}

\date{October 2022}
\def\month{11}  % Insert correct month for camera-ready version
\def\year{2024} % Insert correct year for camera-ready version
\def\openreview{\url{https://openreview.net/forum?id=BXDxwItNqQ}} % Insert correct link to OpenReview for camera-ready version

\begin{document}

\newcommand{\pk}[1]{{\textcolor{black}{#1}}}
\newcommand{\s}[1]{{\textcolor{black}{#1}}}
\newcommand{\tp}[1]{{\textcolor{black}{#1}}}
\newcommand{\edit}[1]{{\textcolor{black}{#1}}}
\newcommand{\edi}[1]{{\textcolor{blue}{#1}}}
\newcommand{\xmark}{\ding{55}}%

\maketitle

\begin{abstract}
Compositional learning, mastering the ability to combine basic concepts and construct more intricate ones, is crucial for human cognition, especially in human language comprehension and visual perception. This notion is tightly connected to generalization over unobserved situations. Despite its integral role in intelligence, there is a lack of systematic theoretical and experimental research methodologies, making it difficult to analyze the compositional learning abilities of computational models. In this paper, we survey the literature on compositional learning of AI models and the connections made to cognitive studies. We identify abstract concepts of compositionality in cognitive and linguistic studies and connect these to the computational challenges faced by language and vision models in compositional reasoning. We overview the formal definitions, tasks, evaluation benchmarks, various computational models, and theoretical findings. Our primary focus is on linguistic benchmarks and combining language and vision, though there is a large amount of research on compositional concept learning in the computer vision community alone. We cover modern studies on large language models to provide a deeper understanding of the cutting-edge compositional capabilities exhibited by state-of-the-art AI models and pinpoint important directions for future research.

\end{abstract}

%%%%%%%%%%%%%%%%%%%%%%%%%%%%%%%%%
%
%      Introduction
%
%%%%%%%%%%%%%%%%%%%%%%%%%%%%%%%%
\section{Introduction} \label{intro}

% What is Compositional Learning
The compositional learning and reasoning of an intelligent agent refers to the ability to understand and manipulate complex structures by decomposing them into simpler parts and composing parts to form new complex concepts with a coherent understanding. This ability is a key factor in generalizing learning to unobserved situations~\citep{hupkes2023stateoftheart}. Compositional learning in intelligent systems is cognitively motivated since humans learn compositionally~\citep{lake2019human}. Researchers have examined this phenomenon from cognitive, linguistic, and psychological perspectives~\citep{doi:10.1126/science.3629243, annurev:/content/journals/10.1146/annurev-psych-122216-011829}. 

% Compositionality, Language as its origin & vision, with examples
The formal notion of compositionality originated from natural language and semantics, with various theories and arguments that elaborate on this concept. The principle of compositionality~\citep{Partee2004-PARCIF, JANSSEN1997417, Montague1974-MONFPS} is defined as \textit{``The meaning of a whole is a function of the meanings of the parts and of the way they are syntactically combined''} with three general methods- new meanings, new basic parts, and new constructions. One of the earliest formalizations of compositionality was grounded in grammar trees when cognitive scientists proposed an \textit{information processing} approach to create a model of the mind~\citep{sep-cognitive-science}. The birth of modern cognitive science happened following the proposal of phrase structure and transformational grammar~\citep{chomsky2014aspects}. Compositional understanding of linguistic constructs has multiple aspects~\citep{marelli-etal-2014-sick,li-etal-2021-compositional}. For example, a nesting description such as ``The black tall woman on the left of the car'' conveys the intersection of multiple adjectives and spatial relations. Thus,  the composition is defined as the intersection of multiple concepts. However, there are cases in which the direct intersection is not applicable, and the meaning should be inferred from the concepts in the global context, such as recognizing the sentiment of the following sentence ``The pizza is so good, I hate this place!'' Despite the natural language being a prominent manifestation of compositionality, this can be expanded to other areas of human intelligence, such as vision~\citep{saffran}. The same notion of intersection, as well as part-whole compositions, is essential for visual intelligence. 

% Why is compositional learning important
Compositional learning is important in complex tasks where high-level goals must be decomposed into smaller subgoals and plans, for example, when instructing an agent to navigate from one point to another~\citep{Schmidhuber1990TowardsCL}. From the computational modeling perspective, traditionally formal grammars have been the means to address the compositional understanding of various modalities primarily in language and extended to vision~\citep{NIPS2011_6faa8040, Hong_2021_ICCV}. While symbolic models inherently address compositional structures, using them alone, that is, parsing raw and noisy data into a structure with manually designed grammars will be brittle in real-world situations. Our study focuses on data-driven approaches based on artificial neural networks and the combined paradigms of neurosymbolic techniques (e.g. see~\cite{rajaby-faghihi-etal-2021-domiknows,premsri2024neurosymbolictrainingreasoningspatial}). Several studies provide both experimental and theoretical analyses, indicating the competitiveness of the neural models in expressing compositional structures such as context-free grammars~\citep{SIEGELMANN1995132, shi2022learning}. Their expressive power added to the robustness in dealing with noisy data makes the neural techniques applicable to realistic situations. 

\begin{figure*}[hbt!]
    \centering
    \includegraphics[width=0.98\textwidth]{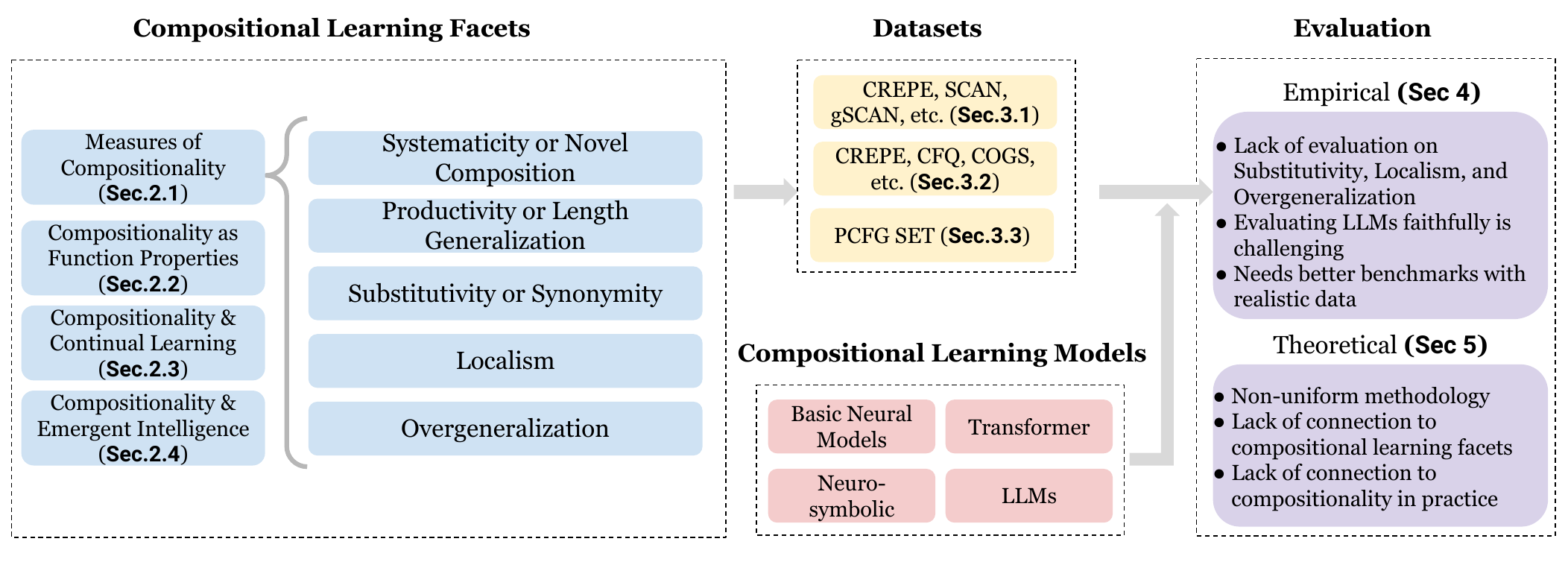}
    \caption{Outline of covered concepts in this survey, related to the structure of the paper.
    We structure our study of compositional learning by dividing it into four parts of compositional learning facets, datasets, compositional learning models, and evaluation methods from both empirical and theoretical perspectives. The topics have the respective sections associated with them. The main areas of required research and future direction are included in the descriptions in the Evaluation boxes, which are further discussed in Section \ref{discussion}.
    }
    \label{fig:overview}
\end{figure*}

{In this paper, we examine multiple aspects of compositional learning, including compositional learning facets, datasets, computational models, and evaluation paradigms in both theory and practice. Figure~\ref{fig:overview} shows the scope and structure of our survey, covering four main topics, that is, Compositional Learning Facets, Datasets, Compositional Learning Models, and Evaluation. 
\textbf{For Compositional Learning Facets}, we overview the different measures of compositionality rooted in cognitive science that define abstract tasks for compositionality and connect them to the other adjacent topics such as continual learning and emergent intelligence. \textbf{For Datasets}, we cover existing benchmarks proposed in the AI community. They help bridge the gap between interdisciplinary theoretical definitions and the design of better evaluation benchmarks to pinpoint model capabilities.
\textbf{For Compositional Learning Models}, we overview the architectures that aim to address compositional learning, divided into categories of basic neural architectures, large language models, and customized architectures, including neurosymbolic models. These models are mostly evaluated empirically using conventional benchmarks, while fewer studies conduct theoretical analyses.  We cover both types of evaluations of various models when available in the existing literature. 
\textbf{For Evaluation}, we overview two evaluation approaches- theoretical and empirical. The theoretical evaluations examine various computational models in a mathematical framework, investigating their expressivity, and capacity for compositional learning, and analyze the generalizations to unobserved situations by computing the error bounds.  Empirical evaluations include experimental results on benchmarks set by datasets and tasks created primarily to highlight the core challenges of compositional learning for language and vision understanding.  Such results often report performance measures on the tasks designed to test the cognitive aspects of compositionality.}

Overall, the cognitive aspects lay the foundation of the concept of compositionality and define the different abstractions associated with it and the tasks that are designed accordingly. The empirical evaluations use these tasks to evaluate compositionality using experimental performance. However, some studies examine the mathematical analysis and functional form of the models independent from the datasets. Finally, both these evaluations are used to develop models that are capable of compositional learning.

%\s{Something explaining the color coding of the figure here or in discussion perhaps?}

% TODO: Fix figure placement, Color code explanation?

%%%%%%%%%%%%%%%%%%%%%%%%%%%%%%%%%%%%%%%%%%%%%%%%%%%%%
%
%     Compositional Learning Facets
%
%%%%%%%%%%%%%%%%%%%%%%%%%%%%%%%%%%%%%%%%%%%%%%%%%%%%%
\section{Compositional Learning Facets} \label{facets}

Compositionality is one important aspect of generalization as a whole \citep{hupkes2023stateoftheart}. Cognitive Science and Linguistics literature have identified broad categorizations of tasks that define compositionality and can be used to evaluate the compositional reasoning of models. The foundations of human natural language lie in compositionality. A commonly used task categorization, derived from reformulated theoretically grounded tests from~\citet{hupkes2020compositionality}, defines five main metrics of compositionality: systematicity, productivity, substitutivity, localism, and overgeneralization~\citep{https://doi.org/10.48550/arxiv.2108.05885}. 

\subsection{Measures of Compositionality}
\citet{hupkes2020compositionality} introduces theoretically grounded tests for compositionality of models based on different interpretations of compositionality~\citep{FODOR19883, 1056813}. {Some of these tasks existed in earlier research under different terms, such as productivity, which builds on a substantial body of prior research on length generalization.} These tasks are becoming widely accepted tasks for compositional learning, and there are current datasets that use them for their evaluation splits. We describe these measures in detail below.

\subsubsection*{Systematicity or Novel Composition}

Systematicity is one of the most commonly used notions of compositionality in evaluating the performance of computational models. It has been defined as the ability to systematically recombine known parts and rules \citep{https://doi.org/10.48550/arxiv.2108.05885}. It derives directly from the commonly accepted definition of composition, which is the formation of compound expressions as a function of simpler ones~\citep{Partee2004-PARCIF}. Systematicity is a standard concept in cognitive science research on building cognitive architectures that try to formalize the human thinking process~\citep{FODOR19883}. The ability to syntactically combine known elements to form new or ``unseen'' expressions is an integral test when evaluating a model's ability to reason compositionally. It is also called Distribution-Based Compositionality Assessment \citep{keysers2020measuring}, where two principles are defined: one is to make sure the distribution of atoms is similar in both training and test sets, while another is to ensure the distribution of compounds is different. For example, if ``red'' and ``car'' are two separately learned concepts, the model should be able to accurately utilize the unseen concept of ``red car.''

\subsubsection*{Productivity or Length Generalization}

Another commonly explored test for compositionality is length generalization or productivity, as defined in \citet{hupkes2020compositionality}. In this evaluation, models are tested on their performance with expressions or sequences that are longer than training data. Longer input sequences may be recursive or nested versions of seen phrases in the case of natural language inputs~\citep{kim2020cogs}. For example, if the model has seen ``A and B,'' it should be able to understand ``A and B and C.''

\subsubsection*{Substitutivity or Synonymity}

Substitutivity is another form of evaluation defined in~\citet{https://doi.org/10.48550/arxiv.2108.05885}, which concerns the model performance on the introduction of synonyms in expressions. For example, testing a model on the translation of the same sentence, \pk{when} switching between synonymous words such as donut to doughnut or aubergine to eggplant \pk{the translation would not change}. \s{This is one of the less explored axioms of compositionality.}%\edit{I am not very clear about this, i remember we have discussed this but can you add a bit of more clarification here @Sania?}

\subsubsection*{Localism}

Another nuance of compositionality is the notion of global versus local composition. According to the principle of compositionality, the locality of the composition operator can vary. 
The meaning of a complex expression can depend solely on the meaning of its immediate parts (local composition) or the global structure of the context. Localism can be tested by analyzing the meaning a model assigns to a standalone compound versus when that compound is part of a larger expression. 
For example, sentences X and Y have the same truth values, but if the same context is added, their composition with the new context might lead to different truth values. For instance, when we add the context, ``Peter thinks'' and obtain ``Peter thinks X'' and ``Peter thinks Y,'' these two new sentences can have different truth values~\citep{hupkes2020compositionality, carnap1947meaning}.  The local interpretation of compositionality says that these new phrases will have the same truth values which might not be the case anymore as Peter might be aware of X and not Y. In other words, considering the phrase that X and Y are a part of, changes the meaning. \s{This is another one of the less explored axioms of compositionality.}

\subsubsection*{Overgeneralization}

Overgeneralization, as defined in \citet{https://doi.org/10.48550/arxiv.2108.05885}, evaluates how much a model prefers an exception versus a rule. The term is originally used in language acquisition literature, also known as overregularization. A well-known example of this is the past-tense debate~\citep{eadfa2ba-d727-3995-86be-56ba36e587c4} in language, \s{which is about the rule that English past-tense verbs can be formed by appending -ed to the
stem of the verb in most cases while there are some exceptions.} %This is another uncommon evaluation task defined for compositionality. 
This property can be evaluated by testing a model on exceptions of a usual rule in the training data and seeing if the model has over-fitted the training samples~\citep{hupkes2020compositionality}. 
Another example of this task is translating idioms where the meaning of sentences is ``exceptions'' to usual rules. \s{For example, when translating the idiom ``it's raining cats and dogs'', a literal translation does not make sense as the phrase is an exception and has a specific meaning different from the usual literal interpretation.} In this scenario, a model can achieve better performance by considering sentences in a global sense -that is looking at the bigger picture, such as context from placement in a compound- instead of trying to evaluate the meaning locally, such as by isolating the phrase. \s{This is yet another less evaluated axiom of compositionality.}

\subsection{Compositionality as Function Properties}
In~\citet{ram2024makes}, compositional functions are defined with several components with a \textit{computation directed acyclic graph} (\textbf{cDAG}) at the core. Their formal definitions facilitate the evaluation of compositional properties of the learning models (i.e. compositional functions). They relate their defined structure to the learning models' expressivity and sample complexity.
In this system, \textbf{Systematicity} can be thought of as the expressivity of a compositional function as a low entropy program (for example, decision tree) versus a high entropy program (for example, transformer).
\textbf{Productivity}, in simple terms, can be interpreted as whether a compositional function is recursive.
\textbf{Substitutivity} tests whether a compositional function respects important abstractions and can be factored over them.
\textbf{Localism} measures the stability of a compositional function against local changes, where the structure of the function's elements affects the importance of the level of locality respected.
\textbf{Overgeneralization} is the extent of compression of a function, where a function might have a general rule but have exceptions to those in special cases. 

\subsection{Compositionality and Continual Learning}
\s{Compositionality is an important aspect of continual learning, also known as lifelong learning~\citep{mendez2021lifelonglearningcompositionalstructures}. Continual learning~\citep{10444954} is the concept of learning new tasks while retaining knowledge from previously learned tasks. In continual learning, compositionality is particularly crucial to prevent catastrophic forgetting, where earlier tasks are forgotten over time due to learning of new tasks~\citep{NEURIPS2023_6a42b45a}. 
Since knowledge about novel tasks is compositional, the existing information can be combined in novel ways and used for future tasks.  This enables forward transfer of knowledge rather than catastrophic forgetting~\citep{mendez2023reusecomposeknowledgelifetime}. Both compositional and continual learning share the key challenge of finding reusable knowledge and further connecting those for transfer learning and dealing with complex unobserved situations.}
%Compositional learning also enables connecting older concepts to the new ones by compositionally forward transfer of knowledge~\citep{NEURIPS2023_6a42b45a}. 
\subsection{Compositionality and Emergent Intelligence}
\tp{
The term emergent behavior has been used across various science-related fields, rooted in ``More Is Different'' by Nobel Prize-winning physicist P.W. Anderson~\citep{doi:10.1126/science.177.4047.393}. 
Its introduction to the language modeling community, specifically in the context of the large language models, begins with~\citet{wei2022emergentabilitieslargelanguage}. 
They defined emergent ability as the ability that appears only in large models and is not observed in any smaller models.
The emergent abilities demonstrated in their study include performing unseen tasks by following instructions~\citep{ouyang2022traininglanguagemodelsfollow} and demonstrating multi-hop reasoning skills through Chain-of-Thought prompting~\citep{wei2023chainofthought}. 
These abilities reflect the model's capacity for generalizing to unobserved situations, which can further extend to the model's compositional learning ability. 
Therefore, the compositional learning ability of the models can be associated with the emergent intelligence of language models. In the same paper of~\citet{wei2022emergentabilitieslargelanguage}, creating new compositional learning benchmarks is proposed as one direction for evaluating and understanding LLMs' emergent abilities.}

%%%%%%%%%%%%%%%%%%%%%%%%%%%%%%%%%%%%%%%%%%%%%%%%%%%%%
%
%     Datasets grouped by abstract tasks
%
%%%%%%%%%%%%%%%%%%%%%%%%%%%%%%%%%%%%%%%%%%%%%%%%%%%%%
\vspace{-1.5mm}
\section{Abstract Tasks and Datasets} \label{datasets}

In this section, we categorize the existing datasets proposed for the evaluation of compositional learning. Our categorization is based on the type of compositionality facet explained in Section~\ref{facets}. Table~\ref{tab:datasets} points to a list of important datasets we have surveyed. 
\s{In general, there are more common evaluation benchmarks for Systematicity and Productivity. Systematicity focuses on the novel composition of seen atomic concepts and there are several benchmarks established for its evaluation, although some of those works do not explicitly use the term systematicity. The productivity measure was often referred to as length generalization before the term became commonly used. However, despite the abundance of datasets for these tasks, most rely on synthetic data, which poses a risk to their reliability in capturing the variation and complexity of real-world problems. While some datasets in the computer vision community for compositional learning utilize realistic images, they address fewer aspects of compositionality (e.g. object-attribute combination) compared to synthesized linguistic corpora designed for the same purpose.}
We describe some of the existing datasets and tasks in this section.

% TODO: Order them chronologically

\subsection{Systematicity or Novel Composition} \label{systematicity}

{\textbf{CREPE.}} This is a Compositional REPresentation Evaluation benchmark (CREPE)~\citep{ma2023crepe}.
The dataset is synthesized and includes multiple splits, one of which relates to Systematicity. The main task setting is that, given an image, the model needs to identify an appropriate text caption describing it among multiple given choices.  This systematicity challenge tests whether the model can systematically generate new combinations of seen atomic concepts during training. For example, ``Crepe on a skillet'' is never observed in the training while Crepe and skillet are observed separately in different contexts. 

\noindent
\textbf{SCAN.}
The task is to navigate in a two-dimensional grid world based on natural language instructions. It is the \textit{Simplified} version of the \textit{CommAI Navigation }tasks (SCAN)~\citep{lake2018generalization, mikolov2016roadmap}. One of the proposed experiments in SCAN evaluates the model's compositional generalization across primitive commands. Specific compounds are excluded from training where the model has seen the primitives and similar compound structures. Then, these unseen compounds are tested during testing. 

\noindent
\textbf{gSCAN.}
The task is to navigate in a two-dimensional grid world based on natural language instructions~\citep{https://doi.org/10.48550/arxiv.2003.05161}, which is a grounded version of SCAN (gSCAN). It includes 9 splits A-I. Categories B to H present tasks that form tests for systematicity (B,C- Novel Composition of Object Properties, D- Novel Direction, E- Novel Contextual References, F- Novel composition of actions and arguments, G,H- Novel Adverbs). Each split focuses on some form of novel composition of known concepts. 

\noindent
\textbf{PCFG SET.}
PCFG SET stands for Probabilistic Context Free Grammar String Edit Task. It is an artificial translation task where sequences produced by probabilistic context-free grammar need to be translated into sequences representing their meaning \citep{hupkes2020compositionality}. The output sequences can be constructed recursively using specified string edit operations applied to the input sequence, e.g., the input ``repeat ABC'' will be mapped to the sequence ``ABCABC.'' The systematicity test uses a combination of words $a$ and $b$ in the input where the model has seen $a$ but not $b$ and vice versa. However, the combination of $a$ and $b$ is plausible in the corpus. 

\noindent
\textbf{COGS.} 
This dataset is designed for the Compositional Generalization Challenge (COGS)~\citep{kim2020cogs}. The task is a kind of semantic parsing based on a fragment of English where the models need to determine a formal meaning representation of the input English text. There are 4 categories in COGS that test some form of systematicity, including Novel Combination of Familiar Primitives and Grammatical Roles, Novel Combination of Modified Phrases and Grammatical Roles, Verb Argument Structure Alteration, and Verb Class.

\noindent
\textbf{ReaSCAN.}
This extension of gSCAN overcomes some of its limitations by requiring compositional language interpretation and reasoning about entities and relations~\citep{wu2021reascan}. The challenges for systematicity are Category A, which tests novel object attribute combinations such as novel color modifier, color attribute, and size attribute, which is adapted from gSCAN~\citep{https://doi.org/10.48550/arxiv.2003.05161}. Category B tests unseen co-occurrences of objects and relations, which is unique to ReaSCAN.

\noindent
\textbf{SQOOP.} 
This is a dataset of Spatial Queries On Object Pairs (SQOOP). \pk{It is a minimalistic visual question answering, with yes-no answers. The input is an image and a question based on spatial reasoning~\citep{bahdanau2019systematic}.}
Models are tested for answering questions on all possible object pairs after being trained on only a subset. Questions are of the form X R Y (X and Y are objects while R is a relation) where training sets are generated by controlled sampling of X and Y objects.

\noindent
\textbf{CLUTRR.} 
This dataset is on Compositional Language Understanding and Text-based Relational Reasoning (CLUTRR). The task is, given a natural language short story, to answer questions on kinship relations that can be inferred from story~\citep{sinha2019clutrr}. Models are tested on combinations of held-out reasoning rules that are unseen during training. Thus, it tests systematic generalization capability or systematicity.

\noindent
\textbf{KiloGram.}
The task is a reference game task where given a textual description, the model has to select the appropriate image from a set of images. These images are tangrams, and the dataset has rich annotations of these images in two splits, FULL and DENSE, which have varying numbers of annotations, hence the name Kilo Tangram KiloGram)~\citep{ji2022abstract}. There are different variations of this task such as showing parts of the image versus the whole image and making it grayscale versus colored. This dataset is an example of compositionality in vision since the whole tangram image is made out of parts, and the model learns how different parts combine to form different images.

\noindent
\textbf{CompMCTG.} To evaluate the compositional learning of generative language models, 
{Compositional multi-aspect controllable text generation (CompMCTG) benchmark\citep{zhong2024benchmarkingimprovingcompositionalgeneralization} is proposed.
The task is to generate sentences given a set of concepts, including aspects of sentiment, topic, tense, person, and stuff.
The benchmark tests systematicity by evaluating the model's capability to generate sentences with novel or unseen combinations of such concepts. For example, if the model has seen sentences with concepts of (red, car) and (blue, hat), it should be able to generate sentences for (blue, car). 
To perform these evaluations, they split the dataset into in-distribution (I.D.) and compositional, which have no intersection, although recombining elements in the I.D. set can form elements in the compositional set. 
Their evaluation protocol includes three test splits of Hold-Out, Attribute Compound Divergence (ACD), and Few-Shot evaluation. 
%These protocols aim to provide a more comprehensive evaluation by making the task of recombination harder. 
It is based on four popular textual corpora- Amazon Reviews~\citep{10.1145/2872427.2883037}, a combination of IMDB, OpeNER, SenTube: Mixture~\citep{liu-etal-2022-multi-attribute}, YELP~\citep{yelp2014dataset}, and Fyelp~\citep{lample2018multipleattribute}. }

\noindent
\s{\textbf{MIT-States.} In this evaluation benchmark~\citep{StatesAndTransformations}, the general task is to explain a collection of images in terms of the novel composition of primitive states and transformations applied to objects. It includes three different tasks: discovering relevant transformations (such as slicing, wilting), parsing states (such as sliced, wilted), and finding smooth transitions. It tests systematicity in image concepts such as being able to identify ``sliced tomato'' versus ``whole tomato'' and generalize it to unobserved compositions such as ``sliced apple'' versus ``whole apple.'' There exists other similar works~\citep{6909426, Yu_2017, 9010671, xu2023metarevisionmetalearningretrievalvisually, bao2024promptinglanguageinformeddistributioncompositional,xu2023gipcolgraphinjectedsoftprompting} in the vision community that evaluate such compositional concept learning.}

\noindent
\s{\textbf{Skill-Mix.} There is no commonly used benchmark for the evaluation of compositional learning abilities of LLMs. A very recent benchmark, called Skill-Mix~\citep{yu2023skillmixflexibleexpandablefamily} covers some aspects of compositional evaluation for generative models. The task expects the models to generate text by combining various skills and imposing some constraints on the generated text. While the notion of compositionality is not highlighted in the paper, we categorize their tasks under systematicity. }

\subsection{Productivity or Length Generalization}

\textbf{CREPE.}
The productivity split of CREPE~\citep{ma2023crepe} evaluates if a model can perform the trained task of longer sets of expressions. In this task, there are variations of complexity for n-subgraphs with  $n\in \{4,5,....12\}$ and variations in the type of hard negatives used in the generation of text options. There are three types of hard negatives used- atomic hard negatives, swap hard negatives, and negation hard negatives. 

\noindent
\textbf{PCFG SET.}
For the Productivity test~\citep{hupkes2020compositionality}, the data is split based on sequence lengths. The model is tested on sequences longer than the ones observed during training. For example, in a grammar context, if the model has been trained on the syntax entity-relation-entity, it will be tested on a longer, nested version of this concept, entity-relation-(entity-relation-entity). 

\noindent
\textbf{CFQ.}
This is a dataset of Compositional Freebase Questions (CFQ). It is a natural language question-answering task ~\citep{keysers2020measuring}, focusing on semantic parsing, with a SPARQL query against the Freebase knowledge base. Questions are generated at varying levels of complexity. Various splits are available based on input length, output length, input pattern, or output pattern. All these splits aim to maximize compound divergence while minimizing atom divergence. This task appears to test both systematicity and productivity to some extent, although not explicitly. However, it cannot identify specific areas where a model's compositional behavior may be deficient.

\noindent
\textbf{COGS.}
\pk{While some splits of this dataset are explained in Section \ref{systematicity}, it also contains a split for length generalization. Category 3 (i.e. Deeper Recursion),} one of the splits of this dataset, ~\citep{kim2020cogs} is a test for length generalization by increasing the length of the input sequence recursively during testing. Input sequences are generated using nesting of phrases that are longer than those seen during training. 

\subsection{Other Generalization Criteria}

To the best of our knowledge, PCFG SET~\citep{hupkes2020compositionality} is the only benchmark that evaluates the other three additional criteria. The \textbf{Substitutivity} or synonymity test uses an input sequence with an atomic unit replaced by a synonymous atomic unit to evaluate how the model prediction changes. \textbf{Localism} is tested by using input sequences composed of smaller sequences A and B. The model is used to translate the full sequences first and then forced to process A and B separately. The outputs of these two experiments are compared to evaluate how local versus global the model is in its compositional reasoning. \textbf{Overgeneralization} test evaluates the model's results on input sequences that do not conform to the general rules of the dataset, that is, input sequences that are exceptions to the dataset rules. For example, the acquisition of past-tense forms such as the common ``ed'' ending (open-opened) versus more uncommon forms such as break to broke. 

% TODO: Rewrite caption for abbreviations?

\begin{table*}[ht!]
    \begin{adjustbox}{width=\columnwidth,center}
    \begin{tabular}{| c || c | c || c | c | c | c | c || c |}
    \hline
     Name & Text & MM. & Sys. & Prod. & Subst. & Loc. & Overgen. & References \\
     \hline
     PCFG SET & \checkmark & \xmark & \checkmark & \checkmark & \checkmark & \checkmark & \checkmark & ~\cite{csordas2022devil} \\
     \hline
     CLUTRR & \checkmark & \xmark & \checkmark & \xmark & \xmark & \xmark & \xmark & 
     \begin{tabular}{@{}c@{}}
     ~\cite{gontier2020measuring}, ~\cite{minervini2020learning} \\
     ~\cite{sinha2019clutrr} \end{tabular} \\
     \hline
     SQOOP & \checkmark & \checkmark & \checkmark & \xmark & \xmark & \xmark & \xmark & ~\cite{damario2022modular}, \cite{bahdanau2019systematic} \\
     \hline
      CFQ & \checkmark & \xmark & \xmark & \checkmark & \xmark & \xmark & \xmark & 
      \begin{tabular}{@{}c@{}}~\cite{furrer2021compositional},~\cite{herzig2021unlocking}, \\ ~\cite{liu2021learning},~\cite{cao2022kqa}\end{tabular}  \\
     \hline
     SCAN & \checkmark & \xmark & \checkmark & \checkmark & \xmark & \xmark & \xmark & \begin{tabular}{@{}c@{}}
     ~\cite{korrel2019transcoding}, ~\cite{nye2020learning}, \\
     ~\cite{dessi2019cnns} \end{tabular} \\
     \hline
     COGS & \checkmark & \xmark & \checkmark & \checkmark & \xmark & \xmark & \xmark & 
     \begin{tabular}{@{}c@{}}
          ~\cite{8953472},~\cite{wu2023recogs},  \\
          ~\cite{klinger2024compositional}
     \end{tabular} \\
     \hline
     gSCAN & \checkmark & \checkmark & \checkmark & \checkmark & \xmark & \xmark & \xmark & ~\cite{gao2020systematic}, ~\cite{spilsbury2023improved} \\
     \hline
     ReaSCAN & \checkmark & \checkmark & \checkmark & \checkmark & \xmark & \xmark & \xmark &
          ~\cite{kamali2023syntaxguided} \\
     \hline
     CREPE & \checkmark & \checkmark & \checkmark & \checkmark & \xmark & \xmark & \xmark & ~\cite{lin2023revisiting},~\cite{singh2023coarsetofine} \\
     \hline
     KiloGram & \checkmark & \checkmark & \checkmark & \xmark & \xmark & \xmark & \xmark & 
     \begin{tabular}{@{}c@{}}
          ~\cite{kojima2023joint},~\cite{gui2023training} \\
          \cite{ji2022abstract}
     \end{tabular} \\
     \hline
     CLEVR & \checkmark & \checkmark & \checkmark & \xmark & \xmark & \xmark & \xmark & 
     \begin{tabular}{@{}c@{}}~\cite{bahdanau2020closure}, \cite{johnson2016clevr},\\~\cite{niemeyer2021giraffe} \end{tabular} \\
     \hline
     MIT-States & \xmark & \checkmark & \checkmark & \xmark & \xmark & \xmark & \xmark & 
     ~\cite{xu2023gipcolgraphinjectedsoftprompting}, \cite{naeem2021learninggraphembeddingscompositional} \\
     \hline
     
    \end{tabular}
    \end{adjustbox}
    \caption{A summary of datasets and the compositional aspects they address with references of relevant papers on compositionality using these datasets. MM: multi-modal, sys: systematicity, prod: productivity, subst: substitutivity, loc: localism, overgen: overgeneralization}
    \label{tab:datasets}
\end{table*}

%%%%%%%%%%%%%%%%%%%%%%%%%%%%%%%%%%%%%%%%%%%%%%%%%%%%%
%
%     Models
%
%%%%%%%%%%%%%%%%%%%%%%%%%%%%%%%%%%%%%%%%%%%%%%%%%%%%%
\vspace{-1.5mm}
\section{Empirical Findings: Compositional Learning Models}

Traditional symbolic AI models naturally support compositional reasoning using classical logic applied to formal language~\citep{szabo2004compositionality}, formal verification~\citep{Giannakopoulou2018}, and more. First-order logic can express objects, their properties, and complex compositional relations. Logical operations like conjunction, disjunction, and implication can express compositional structures on which inference rules are applied, supporting complex compositional reasoning~\citep{10.1007/3-540-45788-7_17}. Another classical symbolic formalism includes grammars~\citep{chomsky2014aspects}, which can express and generate complex compositional structures. 
\pk{However, }dealing with noisy and uncertain data is hard with pure symbolic AI. Moreover, probabilistic augmentations and structured output prediction models have been able to explicitly model structural dependencies and support compositional reasoning based on their learned complex patterns from the data~\citep{pearlBook}. Nevertheless, scalability becomes a challenge for training and inference as the structural dependencies and the number of correlated variables increase. 
%\s{With these brief arguments about the: Given these} 
Given these long-lasting challenges of traditional models of compositionality, current neural models have shown success in both scalability and dealing with noisy and sensory data~\citep{openai2024gpt4}. Especially in modern large language models, complex compositional patterns can be memorized and resemble compositional reasoning. In the rest of this section, we overview the research focused on the development, design, and empirical evaluation of different types of neural models for compositional reasoning. We relate the type of empirical evaluations to the cognitive aspects of compositionality that they are testing. \s{While some of these models utilize tasks and datasets covered in Section \ref{datasets}, others have their own tasks, specific to the problem of their choice.}
%\vspace{-3mm}
\subsection{Basic Neural Models}
In \citet{hupkes2020compositionality}, different neural models are tested on a set of compositional learning tasks. They evaluate Long short-term memory (LSTM) Networks~\citep{hochreiter1997long}, Convolutional Neural Networks (CNN)~\citep{NIPS1989_53c3bce6}, and Transformers for sequence-to-sequence language processing tasks on their proposed PCFG SET tasks including systematicity, productivity, substitutivity, localism and overgeneralization. On average, Transformer outperformed the other two models, but within the two classic neural models, the convolutional model performs better than the LSTM counterpart. In the reported results, two specific architectures, called LSTMS2S and ConvS2S were used. LSTMS2S is a recurrent, bidirectional encoder-decoder model with attention where the encoder and decoder are LSTMs, from the OpenMT-py framework. ConvS2S is a convolution-based sequence-to-sequence model as used in \citet{gehring2017convolutional}. Several other works~\citep{hupkes2018visualisation, zheng-lapata-2021-compositional-generalization, zheng2022disentangled, lake2018generalization} have used similar models to conclude that neural sequence models can exploit recursive compositional structure~\citep{bowman2015treestructured, NIPS2014_2cfd4560} in solving tasks. The related work in compositionality in computer vision indicates similar results. In \citet{klinger2020study}, MLP, CNN, ResNet~\citep{he2015deep}, and relational networks such as WReN~\citep{pmlr-v80-barrett18a} and PrediNet~\citep{shanahan2020explicitly} are evaluated on \s{PCFG SET} substitutivity and productivity tests. Their results indicate that compositional reasoning is challenging for the evaluated models and calls for more sophisticated architectures. 

\subsection{Transformer-based Architectures}\label{transformer_emperical_result}
% General transformers
The compositional capability of large language models is currently a controversial topic.  They have been evaluated on general tasks such as arithmetic, logic, and dynamic programming that are compositional by nature~\cite{nafar-etal-2024-teaching,nafar2024probabilistic}. Some of these evaluation efforts conducted in~\citet{dziri2023faith} concluded that  GPT~\citep{openai2024gpt4} family models, solve these tasks by reducing them to linearized subgraph matching, without developing true compositional reasoning skills. Moreover, it is shown that, asymptotically, they have architectural limitations in solving highly complex compositional tasks with novel patterns due to error propagation of the composition of erroneous building block functions. There is a substantial gap in the performance of Transformers on in-domain and low-complexity compositional examples versus out-of-domain instances. The tested tasks were 1) multi-digit multiplication \citep{book}, 2) Einstein's puzzle, which is a constraint satisfaction problem \citep{https://doi.org/10.1111/j.1467-8640.1993.tb00310.x}, and 3) NP-complete maximum weighted independent set problem \citep{10.5555/1051910}. These tasks are mostly aimed at testing systematicity and productivity. Their results indicate that transformers make predictions on shallow reasoning and memorization of similar subgraph patterns seen during training as opposed to reasoning holistically based on true compositional reasoning. This hypothesis also aligns with the findings presented in \citet{chang2024languagemodelsneedinductive}. They conduct experiments with counting problems, a basic form of length generalization tasks. While transformers can count in observed cases, they dramatically fail to perform out-of-domain for the same task, indicating that transformers rely on memorizing observations. The results on transformers often extend to transformer-based language models. As shown in this work~\citep{anil2022exploringlengthgeneralizationlarge},  fine-tuned transformer-based large language models lack generalization capabilities irrespective of the model size, when tested on other length generalization tasks such as parity and boolean variable assignment. This is due to the transformer's tendency to learn non-sequential patterns that do not apply to longer sequences. However, combining a pretrained LLM's in-context learning ability with scratchpad prompting significantly improves performance on longer sequences. This also implies that despite having access to an infinite data pool, LLMs can potentially learn some tasks better through in-context learning than finetuning, confirmed by theoretical work on LLMs explained in Section~\ref{theory}.

% Influence of architectural decisions in transformers
While a series of research papers focused on evaluating the compositional generalization of Transformers~\citep{dehghani2019universal, Hahn_2020,feng2023revealing} and complex reasoning~\cite{mirzaee-etal-2021-spartqa,mirzaee-kordjamshidi-2022-transfer}, some recent research investigated specific architectural factors that can impact the performance of Transformers on compositional tasks, following the claim that Transformers cannot reason compositionally~\citep{dziri2023faith}. In~\citet{https://doi.org/10.48550/arxiv.2108.04378}, \pk{five configurations were evaluated on several different datasets and benchmarks, by varying five properties} of Transformers including, 1) type of positional encoding, 2) use of copy decoders, 3) model size, 4) weight sharing, and 5) use of intermediate representations for prediction. The employed tasks were Addition, AdditionNegatives, Reversing, Duplication, Cartesian, Intersection, SCAN-length and SCAN-add-jump, PCFG productivity and systematicity, COGS, and CFQ-mcd1. This work concluded that relative positional encodings usually help, but using embeddings is necessary, and merely relative position biases are not sufficient. Tasks like SCAN and CFQ were not affected by positional embeddings. Tasks like Duplication or PCFG benefit from a copy decoder because it can learn a type of symmetry like learning a certain position of the input. As for model size, it was found that for algorithmic tasks, large models did not make a difference. However, for PCFG, large models seemed to outperform their smaller variants. Weight sharing across transformer layers seems to improve accuracy in most tasks. Intermediate representations also improved performance by creating new levels of abstraction that make reasoning easier for solving the end task. Specifically, using intermediate representations achieved state-of-the-art performance on COGS by converting the task from seq2seq to sequence tagging. Using intermediate representation on CFQ, eliminating the need to perform Cartesian products by using triple representations, also showed a significant performance improvement. However, intermediate representations need to be crafted specifically for a task and were only tested on COGS and CFQ. Other works~\citep{ruoss2023randomizedpositionalencodingsboost, kazemnejad2023impactpositionalencodinglength, chang2024languagemodelsneedinductive} also investigate the effect of different positional encodings on the performance of Transformers for specifically length generalization. \citet{kazemnejad2023impactpositionalencodinglength} concludes that commonly used positional encodings such as ALiBi and APE are not well suited for downstream tasks but transformers without any positional encoding (NoPE) outperform other explicit positional encoding methods when testing decoder-only transformers on tasks such as SCAN and copying and reversing.
Despite this, \citet{chang2024languagemodelsneedinductive} presents compelling evidence suggesting that each type of positional encoding demonstrates robustness on unique length generalization tasks. 
Consequently, the performance of Transformers on compositional tasks varies depending on the selected positional encoding.
The authors even suggest integrating multiple positional encodings into Transformers for better compositionality by combining different strengths of those encodings.

% Designing new architecture
Another type of research deviates from evaluating current conventional architectures and instead focuses on designing a new architecture that can compositionally generalize better.
\pk{One of such models,} a multi-modal transformer called GroCoT (Grounded Compositional Transformer)~\citep{sikarwar2022transformers}, was designed and achieved state-of-the-art results on GSRR and ReaSCAN. \pk{Another} multi-modal transformer from~\citet{qiu-etal-2021-systematic} is used as a backbone model with changes to Encoder, Decoder, modified spatial representation, interleaving self-attention, and a modified world state encoding.  This work also showed that a single-layer transformer with a single head can ground and compose novel combinations of visual object attributes. They tested the generalizability on RefEx\pk{, that is on grounding referring expressions and}, their proposed evaluation benchmark based on the target identification subtask of ReaSCAN.
Another \pk{customized architecture} example is adding Pushdown Layers to transformer architecture~\citep{murty2023pushdown} which was recently presented as a replacement for standard self-attention. The recursive structure of natural language is challenging for self-attention layers to capture due to the lack of an explicit recursive-state tracking mechanism, which Pushdown Layers try to overcome. Pushdown layers have a stack tape that helps them model the recursive state of language. This helps Transformer-based language models softly modulate attention over tokens when predicting new ones. Other architectures derive inspiration from cognitive sciences. In the same line of work, RegularGPT~\citep{chi2023transformer}  takes inspiration from working memory. It modifies the Transformer architecture to use weight sharing, adaptive depth, and sliding-dilated attention for better length generalization. When tested on the task of natural language extrapolation, it was found that it captures the local windowed attention patterns, which previous work identified as essential for the task. Additionally, it can efficiently model regular languages such as PARITY.

\subsection{Neurosymbolic Architectures}
% TODO: Decide where this paper goes, connected to the end comment
% \pk{@Sania: A summary of the paper we discussed from Danial, this needs to be integrated, you can still revise a bit. ``The paper presents a compositional network where each token of a parsed command is represented as an RNN unit, structured according to the command's linguistic parse tree. Each unit receives visual observations and attention maps from its child nodes. These attention maps highlight relevant aspects of the visual observations, focusing the network's processing on important details. This setup enables the network to dynamically adjust its focus and interpret novel linguistic combinations effectively and consequently enhancing its ability to handle grounded language tasks in a visually rich environment.''}

A rising trend in cutting-edge research on modeling intelligent systems is neurosymbolic modeling. As the need for general-purpose AI models grows, there is a need for highly compositional models that can reason based on previously trained simpler tasks to do novel and complex ones. Although not explicitly mentioned in this research, they mainly address systematicity and productivity. One approach in this direction is to use natural language explanations to generate formal specifications that explicitly lay out a compositional task in terms of required simpler steps. The formal specifications then are passed to appropriate \textit{engine}s to solve the problem. A prominent vision understanding model that follows this approach is VisProg~\citep{gupta2022visual}.  
VisProg is a modular neurosymbolic model that can solve various compositional visual reasoning tasks given natural language instruction relying merely on the in-context learning of large language models. It produces modular programs in Python to obtain the solution. This approach provides an interpretable reasoning for how the model derives the solution. These modular programs use built-in modules supported by VisProg such as off-the-shelf neural computer vision models, image preprocessing modules, or Python subroutines, and solve complex tasks without any task-specific training. %Examples include replace, background blur, count, crop, etc. VisProg is built on GPT-3 and uses a small number of in-context examples to create modular programs to solve complex tasks without any task-specific training. 
Another example in this line of work is to generate a formal logical specification of the problem from natural language explanations and pass the logical form to a logical reasoner engine~\citep{poesia2023certified}. 
This work uses large language models such as GPT-3 or GPT-3.5 Turbo, for producing ``guides'' to solving complex compositional tasks by breaking those down into smaller steps based on a reasoning chain.
Similar to this, many recent works focused on different prompting strategies that can be used to solve complex compositional tasks with a modular approach. Examples include Decomposed Prompting~\citep{khot2023decomposed}, which uses a modular approach to decompose a complex task into simpler sub-tasks via prompting and pass on these sub-tasks to LLMs that are capable of solving them. This method allows for the optimization of a prompt for a specific sub-task, which can be further decomposed, or replaced with more effective prompts, trained models, or symbolic functions as necessary. A similar approach for mapping to probabilistic logical reasoning is proposed in~\citet{nafar2024probabilistic}. Neurosymbolic modeling has also been used previously in generative models for concept learning~\citep{hofer2021learning} such as in the context of auditory signals for learning evolved combinatorial structure in language.
% TODO: Add a line about explanation. 

Another branch of Neurosymbolic modeling explores leveraging the ability of large language models to do reasoning. This includes casual reasoning with an introduced benchmark, CLadder~\citep{jin2023cladder}.  Their approach is to provide step-by-step structured prompts as a form of a chain-of-thought strategy called CausalCoT. The chain of thought (COT) conveys the formal symbolic representation of the causal reasoning problem.  %A contemporary branch coined Prompt Engineering can improve the reasoning in models with the help of task-specific prompts. Specifically, 
The CoT prompting is influenced by natural language rationales or reasoning processes~\citep{qiao2023reasoning}, which is similar to the use of answer rationales in inducing arithmetic problems~\citep{ling2017program}.

% TODO: is the above better?
% \pk{This can be improved with the help of process optimization like the previous example, as natural language rationales~\cite{ling2017program} significantly influence CoT prompting.: seems grammatical to me, not clear} They can be further divided into different types such as self, ensemble, and iterative optimization~\cite{qiao2023reasoning}.\pk{the whole paragraph needs a revision.}

\textbf{Compositional Neural Architectures.} Over the years, several architectures and theories have been proposed to model compositionality in their design. This aligns with the idea that neural architectures are structurally compositional~\citep{lepori2023break}, meaning they leverage subroutines to break down complex tasks. One of the earlier examples of this type of architecture includes neural modular networks~\citep{andreas2017neural}. Neural modular networks were designed to model the inherent compositionality that exists in linguistic structures. The conceptual modules are built in the neural architecture based on the problem specification. For example, in a visual scene understanding problem, we can place modules for detecting objects, their compositional properties, and relations, which are the main building blocks for abstract reasoning needed for complex scene understanding. 

In a similar line of work~\citep{https://doi.org/10.48550/arxiv.2008.02742}, a network architecture was built that is compositional in nature and makes it possible to interpret what each part of the network learns. It solves tasks in gSCAN, which has agent navigation tasks in a 2D environment following a natural language command. The neural architecture built in this work assembles a \textit{command-specific} network from previously trained modules, modeling the compositional nature of the command (task). 
Later research showed that in Neural Module Networks, it is hard to make the designed modules faithful to expressing the concepts that they are designed for. This is despite the overall network achieving high accuracy for the target task \citep{https://doi.org/10.48550/arxiv.2005.00724}.

\s{Given the importance of compositional learning in lifelong and continual learning, neurosymbolic approaches have been a suitable framework to address continuality through combining program synthesis and neural modeling. One such example is HOUDINI~\citep{10.5555/3327546.3327547}, which is a neurosymbolic framework for lifelong learning of tasks combining perception and procedural reasoning such as counting, summing, and shortest-path computation. They use program synthesis to search over networks described as typed functional programs for the given task, whose parameters are then tuned end-to-end based on stochastic gradient descent. Another example is the Logic-Enhanced Foundation Model (LEFT)~\citep{hsu2023whatsleftconceptgrounding}, a framework designed to learn grounding and reasoning across domains using a differentiable, domain-independent, first-order logic-based program executor. It addresses the lack of generalizability across domains seen in related work, such as VisProg, which we previously discussed, including limitations in generalizing concepts from 2D to 3D images.}

\textbf{Neurocompositional Computing.} Neurosymbolic modeling has been motivated by its connection to \textit{neurocompositional computing}. The term ``neurocompositional computing'' was coined in \citet{smolensky2022neurocompositional}. It defines a type of computing that underlies human cognition as argued in contemporary cognitive science theories in~\citet{cogscismolensky} and incorporates principles of Continuity and Compositionality. The Continuity Principle states that the encoding and processing of information should be continuous, that is, represented by real numbers that vary continuously and can be changed by arbitrarily small amounts. The Compositionality Principle states that larger, more complex structures can be decoded on the basis of smaller, simpler, and familiar building blocks. According to the \textit{Central Paradox of Cognition}, the human brain follows both a continuous neural computing structure and a discrete compositional-structure computer. Following this theory, neurosymbolic models that are both continuous and discrete in architecture seem like the ideal approach to modeling compositional behavior in \pk{computational models}~\citep{Smolensky2022NeurocompositionalCI}.
% TODO: \pk{Mapping natural language to symbolic problem specifications -- we need more citation here.}
% TODO: Decide on this
% \pk{this explanation should move to the beginning it seems to be referring to a very specific arch. maybe we can explain ti better also. }

%%%%%%%%%%%%%%%%%%%%%%%%%%%%%
%
%     Theory
%
%%%%%%%%%%%%%%%%%%%%%%%%%%%%%

\section{Theoretical Findings: Mathematical Formulations of Compositionality} \label{theory}

Theoretical analysis is fundamental for deepening our understanding of the compositionality of learning models. It can reveal intriguing and previously uncovered information that experimental analysis may overlook. 
Many research works have proposed diverse approaches for investigating the compositionality of learning models.
We highlight three different approaches, including a mathematical framework for defining compositionality~\citep{ram2023how}, 
exploring the upper-bounds of expressivity that relate to compositionality~\citep{merrill2023parallelism}, 
and analyzing error-bounds to demonstrate the model's limitations in solving compositional learning  problems~\citep{dziri2023faith}. 
In the rest of this section, we provide a detailed overview of these cases and explain the theoretical results on compositional generalization of classical neural networks, transformers, and modern language models. We also relate the mentioned techniques to aspects of compositionality when applicable.

\subsection{Classical Neural Network} 
\citet{ram2023how} provides a mathematical definition of compositionality for learning models and connects their expressively to computational complexity. 
They frame the existing well-known models, such as variations of RNN and CovNets, with the provided formal definition to explain properties related to their compositional generalization. 
\citet{hewitt-etal-2020-rnns} further investigates the RNN's ability to generate natural language with a certain nesting depth. 
They claim that RNNs with optimal memory and $O(m \log k)$ hidden units can generate a natural language of well-nested brackets of $k$ types and $m$ bounded nesting depth. 
With the rise of LLMs, compositional generalization has recently become more critical. Due to their large-scale parameters and training data, LLMs perform empirically well on many tasks. However, the empirical performance measures are now less reliable, as the high performance on test data can not be interpreted as compositional generalization anymore. This issue is due to the nature of internet-scale training of LLMs and data contamination.
Consequently, there is more urgency for theoretical studies to understand their limitations and measure their reliability in unobserved situations. However, \citet{ahn2023learning} argued that studying the smaller models at the single neuron level potentially leads to a better understanding of the large/deep models' learning behavior, 
{which is related to explaining the Systematicity of the model. }
They also establish a connection between the \textit{Edge of Stability } identified by the learning rate of the gradient descent approach for non-convex optimization and the emergent abilities in learning. 
This result remains limited to the scope of a single neuron and has not yet been extended to large models.

% Transformer
\subsection{Transformers} 
To define the limitations of LLMs, it is essential to investigate the limitations of transformers and their underlying architectural component. In this work~\citep{merrill2023parallelism}, the authors assume a specific transformer type, suggesting that their arithmetic precision is logarithmic in the number of input tokens. 
Based on this assumption, they demonstrate that transformers cannot accurately solve linear equalities or check membership in an arbitrary context-free grammar with empty productions. 
The studies of transformer precision have been explored before in~\citet{dehghani2019universal}. 
They claim that standard transformers have limited precision, implying that they cannot handle an infinite input length. 
{This conclusion indicates the limitations in of the compositionally of the transformers in terms of the Productivity aspect.}
Another notable theoretical investigations focus on the activation functions to explain the limitation of the transformer~\citep{dehghani2019universal,Hahn_2020}. 
\citet{Hahn_2020} analyzes both hard-attention and soft-attention transformers. 
For hard attention, they prove that the transformer ignores most of the input information diagnosed by the specific modifications applied to the input. 
According to their analysis, transformers with hard attention will be unable to solve problems that require processing the entire input, such as PARITY and logical formula problems. 
However, this conclusion contradicts older papers that state transformers are Turing complete~\citep{JMLR:v22:20-302}. 
They utilize the strong assumption that all input information is accessible using hard attention to prove Turing completeness. 
This leads to a different conclusion, stating that the transformer can compute and access the entire internal dense representation. 
\citet{Hahn_2020} also investigate the model's behavior with soft attention.
They illustrate that Transformers struggle with solving long input by demonstrating the influence of input on output substantially drops as the input gets longer. This is a similar conclusion, using a different approach analyzed in an older study of~\citep{dehghani2019universal}. 
{Based on these analyses, they further confirm the lack of productivity aspect of transformer architecture caused by limited training data lengths.}

Despite proving weaknesses of the transformer, \citet{Hahn_2020} claims that the transformer has the potential to solve small input tasks completely.
Two recent works also support this claim. 
The first work provides proof utilizing computation graphs and a theoretical study of error propagation in transformers. 
They claim that the auto-regressive transformer's error reduces as the size of the input decreases~\citep{dziri2023faith}. 
Moreover, they show that transformers reduce problems into multi-step compositional problems to solve larger tasks, {which is strongly related to the Novel Composition of the compositional aspects.}
The second work supports the mentioned claim based on the study of sub-sequential finite-state transducers (SFSTs)~\citep{valvoda-etal-2022-benchmarking}. 
They generate a set of random SFSTs following Montague’s Compositionality theorem to discover the coverage limitation. This limitation is inversely related to the size of the dataset and significantly impacts the probability of a model's successful generalization.

% Large Language Models
\subsection{Large Language Models} \label{llm}
In addition to inconclusive theoretical studies on transformer limitations, there are controversial results on large language models.
The most noteworthy study is on the emerging abilities and capabilities claimed to be unique in the large models.  
\tp{The emerging abilities relate to the generalization to new and complex tasks in LLMs. This kind of ability is also a feature of models' compositional learning ability, allowing them to perform in novel compositional situations~\citep{yu2023skillmixflexibleexpandablefamily}.} 
\tp{Multiple works have shown the existence of emergent abilities of LLMs~\citep{wei2022emergentabilitieslargelanguage}. The recent work of~\citet{arora2023theoryemergencecomplexskills} provides a mathematical framework for identifying complex skills in language models. They use the LLM Scaling Rule to argue that emergent skills are the results of reducing excessive loss. This excessive loss enables the model to learn how to utilize and combine skills from downstream tasks during training. Their claims are based on the assumption that language inherently contains a random mix of complex skills.}
Although several experiments reveal these emerging capabilities, at least two papers disclaim their existence. 
The first group provides a theoretical proof based on a mathematical framework. They illustrate that the emerging ability appears due to the selected evaluation metrics that are nonlinear and discontinuous~\citep{schaeffer2023emergent}. 
They show as an artifact of the evaluation metrics, even simple models such as CNNs can show emerging abilities. 
Therefore, they conclude that emerging abilities may not be a fundamental property of the large models. Moreover, ~\citet {lu2023emergent} provides an extensive empirical study with 1000 experiments on 22 tasks with different LLMs. However, given the inconsistency in some results and the unpredictability of emerging abilities, they do not find any strong evidence of how they emerge.
They associate the performance with in-context learning techniques, memorization, and data contamination. 
However, a recent study presents a positive theoretical analysis of reasoning capabilities by investigating the chain of thought (CoT)~\citep{wei2023chainofthought} and draws a different conclusion. 
They argue that the log-precision transformer can perform fundamental operations such as multiplication and a look-up table. 
Consequently, it can solve linear equations and other reasoning problems if it stores all the input information. 
However, the architecture alone struggles with storing the entire input, as observed in~\citet{dehghani2019universal, merrill2023parallelism}. They show that the model addresses this limitation by repeatedly referring to the input by enabling CoT~\citep{feng2023revealing}. 
Therefore, with the right number of CoT examples, LLMs can overcome the transformer's weakness in solving mathematical reasoning. \tp{This is aligned with previous empirical results of in-context learning discussed in Section~\ref{transformer_emperical_result}.}

\begin{table*}[ht!]
    \begin{adjustbox}{width=\columnwidth,center}
    
    \begin{tabular}{|*{4}{c|}}
      \hline
        & Model Type & Theoretical Analysis & Empirical \\
        \hline
        Basic Neural Models & RNN & \citealt{hewitt-etal-2020-rnns} & \citealt{bowman2015treestructured}  \\
        \cline{2-4}
        & CNN & \xmark &  \citealt{hupkes2020compositionality}\\
        \cline{2-3}
        & LSTM & \citealt{SIEGELMANN1995132} &  \\
        \hline
        Transformer-based & (Customized) Transformers & 
        \begin{tabular}{@{}c@{}}~\citealt{Hahn_2020},~\citealt{perez2018on}, \\ ~\citealt{dehghani2019universal}\end{tabular}  & \citealt{https://doi.org/10.48550/arxiv.2108.04378}  \\
        \cline{2-4}
        % & Pretrained LM & &  \\
        % \cline{2-4}
        & LLM & \citealt{dziri2023faith} &  \citealt{schaeffer2023emergent} \\
        \hline
        Neurosymbolic & Neural Modular Network & \xmark  & ~\citealt{https://doi.org/10.48550/arxiv.2008.02742} \\
        \cline{2-4}
        &  Other Models & \xmark & ~\citealt{gupta2022visual}  \\
        \cline{2-4}
        \hline
        
    \end{tabular}
        \end{adjustbox}
    \caption{Summary of computational models with compositional learning ability from the theoretical perspective and an example from the experimental perspective.}
    \label{tab:models}
\end{table*}

%%%%%%%%%%%%%%%%%%%%%%%%%%%%%
%
%     Discussion
%
%%%%%%%%%%%%%%%%%%%%%%%%%%%%%
\section{Discussion and Future Direction} \label{discussion}
% \pk{We discussed the abstract compositionality tasks and their related benchmarks, we need to discuss how these abstract notions flow in real-world and more general tasks. For example, for systematicity and novel situations in visual reasoning if we can do part-whole and spatial reasoning that will help in downstream applications. Or video understanding and the sequence of events that needs temporal reasoning.  }
% TODO: Is there anything else to add to discussion? Our own suggestions for tests? Tests for substitutivity

There has been a large amount of research on the compositional learning ability of humans from a cognitive perspective~\citep{FODOR19883, NEURIPS2022_d0241a0f}. Researchers in linguists and formal languages have formalized the notion of compositionality since languages have inherent compositional structure~\citep{chomsky2014aspects, Chomsky+2002+115+118}. However, from the AI and machine learning perspective, ideas are borrowed from both cognitive and linguistics, and computational tasks and models are designed focusing on narrow aspects of compositionality~\citep{hupkes2020compositionality}. Our investigation of AI models indicates several challenges regarding the designed tasks, benchmarks, and theoretical frameworks that make the evaluation of computational models problematic. 

Figure \ref{fig:overview} shows the connections between the main topics identified and discussed in this survey. 
Among the five identified types of compositional learning facets, only systematicity and productivity are well-researched and have clear connections to evaluation benchmarks. Given that these are the five main metrics of compositionality, we should aim to expand our evaluation capabilities by developing tests for the other three as well. Empirical evaluations are comparatively more well-studied compared to theoretical analyses. Theoretical evaluations are either lacking or do not follow a consistent methodology. 
There is a lack of connection between the theoretical methods and the cognitive aspects, making these results hard to use to guide better architectural design. For Models, different types of architectures have been designed. However, evaluating LLMs and making fundamental design decisions for compositional generalization present new challenges.
We outline some of these challenges in detail below.

\textbf{Less Explored Facets of Compositionality.} Only Systematicity and Productivity have been well-researched and have established connections to evaluation benchmarks. While the other three were introduced as fundamental types of compositionality, they have received less attention, as they appear to be less commonly occurring aspects of compositionality.
However, in the era of LLMs and the emergent in-context learning, Substitutivity and Localism are potential bottlenecks in the performance of LLMs for attending the appropriate context for solving problems. Moreover,  Overgeneralization can be associated with hallucination and generating unfounded and incorrect information by making up new unrealistic abstractions. While hallucination is a broader concept than overgeneralization, this compositional learning facet can highlight an important aspect to be addressed to prevent hallucination~\citep{huang2023surveyhallucinationlargelanguage}. Therefore, directing attention to the other three types of measures can help establish new formal evaluation benchmarks, contributing to the development of more robust systems and addressing the challenges of the LLMs.

\textbf{Synthetic and Unrealistic Evaluations.} One issue in current evaluations is that controlled and clean tests of compositonality are mostly synthesized~\citep{wu2021reascan, https://doi.org/10.48550/arxiv.2003.05161}. \s{This is evidenced by our examples in Section \ref{datasets}.} Even in rare cases that claim to work with realistic data~\citep{keysers2020measuring}, synthesized questions are used to query knowledge graphs. %\pk{not sure what do you mean by applicability to KGs here, we can discuss}. 
However, more recent studies on language models' evaluation of compositionality focus on more challenging problems such as multi-hop question answering~\citep{press2023measuring, liu2023compositional,okawa2023compositional, mirzaee-etal-2021-spartqa} as well as complex puzzles with combinatorial search solutions or compositional mathematical reasoning~\citep{dziri2023faith}. \s{Although the benchmarks are designed for evaluating specific tasks, the reliance on mostly synthesized data risks the effectiveness of generalization to real-world data, which is often more complex.}
% \s{A lack of realistic data in evaluation can lead to data contamination and poses a challenge in the reliability of evaluation of the abilities LLMs~\citep{sainz2023nlpevaluationtroubleneed}, which is discussed in the following paragraph.}

\textbf{Misalignment of Performance and Compositional Learning (LLM Evaluation Challenge).} The second challenge that mostly applies to LLMs is data contamination. Though the recent research compares language models to the specialized architectures and indicates their outperformance in compositional tasks~\citep{furrer2021compositional},  this result does not necessarily mean these models have better generalizations in recognizing unobserved compositions~\citep{press2023measuring}. 
A major issue with these evaluations on realistic data is the difficulty in disentangling the compositional reasoning from the data contamination~\citep{sainz2023nlpevaluationtroubleneed} and memorization. \s{However, very little recent work has been done to address this issue~\citep{yu2023skillmixflexibleexpandablefamily}.}
% \s{[cite some papers here that try to separate contamination]}
\s{Based on our investigation of theoretical studies in Section \ref{theory}, we identify that the generalization abilities can be an artifact of observing more complex data in larger contexts as well as evaluation metrics as has been pointed out in~\citet{schaeffer2023emergent} and is a pertinent issue that needs to be addressed in new evaluation benchmarks.}

\textbf{Inconsistent Theoretical Methodology.} Given the difficulty in obtaining conclusive empirical studies, theoretical understandings become even more important, nowadays. However, the lack of a well-established and practically informative theoretical framework for investigating the limitations and capabilities of LLMs has been a challenge to a deep understanding of their generalizability. 
\s{According to our studies on the theoretical explanation of transformers in Section \ref{theory}}, 
their compositionality is still under discussion. 
Some results illustrate that transformers possess compositional generalizability based on their ability to solve complex tasks based on smaller subtasks~\citep{feng2023revealing, dziri2023faith}. 
\tp{However, other results based on a different evaluation methodology suggest that the emergence of such abilities, including compositional learning, is associated with the user's choice of evaluation metrics~\citep{schaeffer2023emergent}. Similar lines of study deny the emergence of intelligence and relate the new abilities to in-context learning methods, models memory, and linguistic knowledge~\citep{lu2023emergent}.}
Many research results confirm the limitations in the generalizability of LLMs. 
For example, the building blocks of these models, i.e. transformers, still have severe limitations in comprehending large inputs~\citep{Hahn_2020,dehghani2019universal}. 
\tp{Despite these controversial discussions, there are only relatively few studies on the theoretical analysis of transformer-based models. 
The methodological frameworks for examining the LLMs' capabilities are not standardized.}
Therefore, the compositional capabilities of the current SOTA models require more attention from the research community. This research direction will help towards more consistent and conclusive results on the limitations of the model's generalizability, specifically the compositional generalization. 

\textbf{Cognitive Motivation.} The fundamental capabilities of current AI models have been debated and criticized by scientists in cognitive science and psychology~\citep{bender-koller-2020-climbing,marcus2018deep}. Despite giant leaps of performance progress in modern AI, there are distinct differences between these machines and human intelligence. \s{From our exploration of cognitive aspects of compositional learning in Sections~\ref{intro} and \ref{facets}, we observe that cognitive foundations do not have strong ties to model building yet. }Evaluating different models reveals that they often rely simply on pattern recognition~\citep{Geirhos_2020, dziri2023faith}, instead of a holistic understanding of a problem grounded in reality and situation, \s{as was seen in Section \ref{theory}}. Understanding human intelligence from Cognitive Science literature suggests that we must move beyond current engineering trends to build causal models of the world that support knowledge and understanding. The key ingredients of such human-like rich and efficient learning are compositionality and learning-to-learn~\citep{Lake2017-LAKBMT}.
% \pk{to make claims stronger maybe we should refer to the sections again.} \s{This one feels a little hard to connect with the sections since we do not have a section on just cognitive science really}
\section{Limitations}
Despite the comprehensive nature of the survey and our efforts to cover and connect most research relating to compositional learning, we would like to acknowledge some limitations. 
The scope of this survey covers a broad spectrum of topics and tries to capture both theoretical and experimental frameworks, but there might be some relevant papers that are missed. 
Compositional learning is an interdisciplinary topic across Computer Science, Linguistics, Cognitive Science, etc. Although we have included insights and connections from across these fields, our work has a more in-depth focus on Computer Science literature, especially Natural language processing.
%Additionally, there are diverse interpretations and applications of compositional learning across relevant fields. In this survey, we try to bridge the gap between different variations. 
While we tried to provide the overall picture of the related research and build a coherent story, we might not capture the detailed nuances of each definition and application.

%\section{Limitations} 

\section*{Acknowledgements}
This project is partially supported by the Office of Naval Research (ONR) grant N00014-23-1-2417. Any opinions, findings, conclusions, or recommendations expressed in this material are those of the authors and do not necessarily reflect the views of the Office of Naval Research.
We thank Dr. Tim Klinger and the anonymous reviewers for their constructive feedback, which greatly helped us improve this manuscript. 
 
% TODO: here
% Besides the above challenges, there are some topics of compositionality that are underexplored such as substitutivity or overgeneralism. Paragraph here- Parisa?
% \pk{In the context of natural language, this is a point of debate since replacing an expression with its synonym might change the meaning of the whole. However, in this respect, it might be relevant to look at the paraphrasing task as a test of compositionality. }
% \pk{Replacing a word with its synonym in a way that does not change the semantics of the sentence.}
% Currently, compositionality is divided into five abstract tasks- systematicity, productivity, substitutivity, localism, and overgeneralization. However, there are specific applications of compositionality that are hard to categorize into one specific group. Two such tasks are part-whole composition for spatial or temporal reasoning. 
\bibliographystyle{tmlr}
\bibliography{main}

\begin{thebibliography}{156}
\providecommand{\natexlab}[1]{#1}
\providecommand{\url}[1]{\texttt{#1}}
\expandafter\ifx\csname urlstyle\endcsname\relax
  \providecommand{\doi}[1]{doi: #1}\else
  \providecommand{\doi}{doi: \begingroup \urlstyle{rm}\Url}\fi

\bibitem[Ahn et~al.(2023)Ahn, Bubeck, Chewi, Lee, Suarez, and
  Zhang]{ahn2023learning}
Kwangjun Ahn, Sebastien Bubeck, Sinho Chewi, Yin~Tat Lee, Felipe Suarez, and
  Yi~Zhang.
\newblock Learning threshold neurons via edge of stability.
\newblock In \emph{Thirty-seventh Conference on Neural Information Processing
  Systems}, 2023.
\newblock URL \url{https://openreview.net/forum?id=9cQ6kToLnJ}.

\bibitem[Anderson(1972)]{doi:10.1126/science.177.4047.393}
P.~W. Anderson.
\newblock More is different.
\newblock \emph{Science}, 177\penalty0 (4047):\penalty0 393--396, 1972.
\newblock \doi{10.1126/science.177.4047.393}.
\newblock URL
  \url{https://www.science.org/doi/abs/10.1126/science.177.4047.393}.

\bibitem[Andreas et~al.(2016)Andreas, Rohrbach, Darrell, and
  Klein]{andreas2017neural}
Jacob Andreas, Marcus Rohrbach, Trevor Darrell, and Dan Klein.
\newblock Neural module networks.
\newblock In \emph{2016 IEEE Conference on Computer Vision and Pattern
  Recognition (CVPR)}, pp.\  39--48, 2016.
\newblock \doi{10.1109/CVPR.2016.12}.

\bibitem[Anil et~al.(2022)Anil, Wu, Andreassen, Lewkowycz, Misra, Ramasesh,
  Slone, Gur-Ari, Dyer, and
  Neyshabur]{anil2022exploringlengthgeneralizationlarge}
Cem Anil, Yuhuai Wu, Anders~Johan Andreassen, Aitor Lewkowycz, Vedant Misra,
  Vinay~Venkatesh Ramasesh, Ambrose Slone, Guy Gur-Ari, Ethan Dyer, and Behnam
  Neyshabur.
\newblock Exploring length generalization in large language models.
\newblock In Alice~H. Oh, Alekh Agarwal, Danielle Belgrave, and Kyunghyun Cho
  (eds.), \emph{Advances in Neural Information Processing Systems}, 2022.
\newblock URL \url{https://openreview.net/forum?id=zSkYVeX7bC4}.

\bibitem[Arora \& Goyal(2023)Arora and
  Goyal]{arora2023theoryemergencecomplexskills}
Sanjeev Arora and Anirudh Goyal.
\newblock A theory for emergence of complex skills in language models, 2023.
\newblock URL \url{https://arxiv.org/abs/2307.15936}.

\bibitem[Bahdanau et~al.(2019)Bahdanau, Murty, Noukhovitch, Nguyen, de~Vries,
  and Courville]{bahdanau2019systematic}
Dzmitry Bahdanau, Shikhar Murty, Michael Noukhovitch, Thien~Huu Nguyen, Harm
  de~Vries, and Aaron Courville.
\newblock Systematic generalization: What is required and can it be learned?
\newblock In \emph{International Conference on Learning Representations}, 2019.
\newblock URL \url{https://openreview.net/forum?id=HkezXnA9YX}.

\bibitem[Bahdanau et~al.(2020)Bahdanau, de~Vries, O'Donnell, Murty, Beaudoin,
  Bengio, and Courville]{bahdanau2020closure}
Dzmitry Bahdanau, Harm de~Vries, Timothy~J. O'Donnell, Shikhar Murty, Philippe
  Beaudoin, Yoshua Bengio, and Aaron Courville.
\newblock Closure: Assessing systematic generalization of clevr models, 2020.

\bibitem[Bao et~al.(2024)Bao, Chen, Huang, and
  Kong]{bao2024promptinglanguageinformeddistributioncompositional}
Wentao Bao, Lichang Chen, Heng Huang, and Yu~Kong.
\newblock Prompting language-informed distribution for compositional zero-shot
  learning, 2024.
\newblock URL \url{https://openreview.net/forum?id=LAEd3kHao9}.

\bibitem[Barrett et~al.(2018)Barrett, Hill, Santoro, Morcos, and
  Lillicrap]{pmlr-v80-barrett18a}
David Barrett, Felix Hill, Adam Santoro, Ari Morcos, and Timothy Lillicrap.
\newblock Measuring abstract reasoning in neural networks.
\newblock In Jennifer Dy and Andreas Krause (eds.), \emph{Proceedings of the
  35th International Conference on Machine Learning}, volume~80 of
  \emph{Proceedings of Machine Learning Research}, pp.\  511--520. PMLR, 10--15
  Jul 2018.
\newblock URL \url{https://proceedings.mlr.press/v80/barrett18a.html}.

\bibitem[Bender \& Koller(2020)Bender and Koller]{bender-koller-2020-climbing}
Emily~M. Bender and Alexander Koller.
\newblock Climbing towards {NLU}: {On} meaning, form, and understanding in the
  age of data.
\newblock In Dan Jurafsky, Joyce Chai, Natalie Schluter, and Joel Tetreault
  (eds.), \emph{Proceedings of the 58th Annual Meeting of the Association for
  Computational Linguistics}, pp.\  5185--5198, Online, July 2020. Association
  for Computational Linguistics.
\newblock \doi{10.18653/v1/2020.acl-main.463}.
\newblock URL \url{https://aclanthology.org/2020.acl-main.463}.

\bibitem[Bowman et~al.(2015)Bowman, Manning, and
  Potts]{bowman2015treestructured}
Samuel~R. Bowman, Christopher~D. Manning, and Christopher Potts.
\newblock Tree-structured composition in neural networks without
  tree-structured architectures.
\newblock In \emph{Proceedings of the 2015th International Conference on
  Cognitive Computation: Integrating Neural and Symbolic Approaches - Volume
  1583}, COCO'15, pp.\  37–42, Aachen, DEU, 2015. CEUR-WS.org.

\bibitem[Cao et~al.(2022)Cao, Shi, Pan, Nie, Xiang, Hou, Li, He, and
  Zhang]{cao2022kqa}
Shulin Cao, Jiaxin Shi, Liangming Pan, Lunyiu Nie, Yutong Xiang, Lei Hou,
  Juanzi Li, Bin He, and Hanwang Zhang.
\newblock {KQA} pro: A dataset with explicit compositional programs for complex
  question answering over knowledge base.
\newblock In Smaranda Muresan, Preslav Nakov, and Aline Villavicencio (eds.),
  \emph{Proceedings of the 60th Annual Meeting of the Association for
  Computational Linguistics (Volume 1: Long Papers)}, pp.\  6101--6119, Dublin,
  Ireland, May 2022. Association for Computational Linguistics.
\newblock \doi{10.18653/v1/2022.acl-long.422}.
\newblock URL \url{https://aclanthology.org/2022.acl-long.422}.

\bibitem[Carnap(1947)]{carnap1947meaning}
Rudolf Carnap.
\newblock \emph{Meaning and necessity: A study in semantics and modal logic}.
\newblock University of Chicago Press, 1947.

\bibitem[Chang \& Bisk(2024)Chang and
  Bisk]{chang2024languagemodelsneedinductive}
Yingshan Chang and Yonatan Bisk.
\newblock Language models need inductive biases to count inductively, 2024.
\newblock URL \url{https://arxiv.org/abs/2405.20131}.

\bibitem[Chi et~al.(2023)Chi, Fan, Rudnicky, and Ramadge]{chi2023transformer}
Ta-Chung Chi, Ting-Han Fan, Alexander Rudnicky, and Peter Ramadge.
\newblock Transformer working memory enables regular language reasoning and
  natural language length extrapolation.
\newblock In Houda Bouamor, Juan Pino, and Kalika Bali (eds.), \emph{Findings
  of the Association for Computational Linguistics: EMNLP 2023}, pp.\
  5972--5984, Singapore, December 2023. Association for Computational
  Linguistics.
\newblock \doi{10.18653/v1/2023.findings-emnlp.397}.
\newblock URL \url{https://aclanthology.org/2023.findings-emnlp.397}.

\bibitem[Chomsky(1956)]{1056813}
N.~Chomsky.
\newblock Three models for the description of language.
\newblock \emph{IRE Transactions on Information Theory}, 2\penalty0
  (3):\penalty0 113--124, 1956.
\newblock \doi{10.1109/TIT.1956.1056813}.

\bibitem[Chomsky(1965)]{chomsky2014aspects}
Noam Chomsky.
\newblock \emph{Aspects of the Theory of Syntax}.
\newblock The MIT Press, Cambridge, 1965.
\newblock URL
  \url{http://www.amazon.com/Aspects-Theory-Syntax-Noam-Chomsky/dp/0262530074}.

\bibitem[Chomsky(2002)]{Chomsky+2002+115+118}
Noam Chomsky.
\newblock \emph{Backmatter}, pp.\  115--118.
\newblock De Gruyter Mouton, Berlin, New York, 2002.
\newblock ISBN 9783110218329.
\newblock \doi{doi:10.1515/9783110218329.bm}.
\newblock URL \url{https://doi.org/10.1515/9783110218329.bm}.

\bibitem[Csord{\'a}s et~al.(2021)Csord{\'a}s, Irie, and
  Schmidhuber]{csordas2022devil}
R{\'o}bert Csord{\'a}s, Kazuki Irie, and Juergen Schmidhuber.
\newblock The devil is in the detail: Simple tricks improve systematic
  generalization of transformers.
\newblock In Marie-Francine Moens, Xuanjing Huang, Lucia Specia, and Scott
  Wen-tau Yih (eds.), \emph{Proceedings of the 2021 Conference on Empirical
  Methods in Natural Language Processing}, pp.\  619--634, Online and Punta
  Cana, Dominican Republic, November 2021. Association for Computational
  Linguistics.
\newblock \doi{10.18653/v1/2021.emnlp-main.49}.
\newblock URL \url{https://aclanthology.org/2021.emnlp-main.49}.

\bibitem[D'Amario et~al.(2021)D'Amario, Sasaki, and Boix]{damario2022modular}
Vanessa D'Amario, Tomotake Sasaki, and Xavier Boix.
\newblock How modular should neural module networks be for systematic
  generalization?
\newblock In A.~Beygelzimer, Y.~Dauphin, P.~Liang, and J.~Wortman Vaughan
  (eds.), \emph{Advances in Neural Information Processing Systems}, 2021.
\newblock URL \url{https://openreview.net/forum?id=dwY40cSK-dt}.

\bibitem[Dankers et~al.(2021)Dankers, Bruni, and
  Hupkes]{https://doi.org/10.48550/arxiv.2108.05885}
Verna Dankers, Elia Bruni, and Dieuwke Hupkes.
\newblock The paradox of the compositionality of natural language: a neural
  machine translation case study, 2021.
\newblock URL \url{https://arxiv.org/abs/2108.05885}.

\bibitem[Dehghani et~al.(2019)Dehghani, Gouws, Vinyals, Uszkoreit, and
  Kaiser]{dehghani2019universal}
Mostafa Dehghani, Stephan Gouws, Oriol Vinyals, Jakob Uszkoreit, and Lukasz
  Kaiser.
\newblock Universal transformers.
\newblock In \emph{International Conference on Learning Representations}, 2019.
\newblock URL \url{https://openreview.net/forum?id=HyzdRiR9Y7}.

\bibitem[Dess{\`\i} \& Baroni(2019)Dess{\`\i} and Baroni]{dessi2019cnns}
Roberto Dess{\`\i} and Marco Baroni.
\newblock {CNN}s found to jump around more skillfully than {RNN}s:
  Compositional generalization in seq2seq convolutional networks.
\newblock In Anna Korhonen, David Traum, and Llu{\'\i}s M{\`a}rquez (eds.),
  \emph{Proceedings of the 57th Annual Meeting of the Association for
  Computational Linguistics}, pp.\  3919--3923, Florence, Italy, July 2019.
  Association for Computational Linguistics.
\newblock \doi{10.18653/v1/P19-1381}.
\newblock URL \url{https://aclanthology.org/P19-1381}.

\bibitem[Dziri et~al.(2023)Dziri, Lu, Sclar, Li, Jiang, Lin, Welleck, West,
  Bhagavatula, Le~Bras, Hwang, Sanyal, Ren, Ettinger, Harchaoui, and
  Choi]{dziri2023faith}
Nouha Dziri, Ximing Lu, Melanie Sclar, Xiang~(Lorraine) Li, Liwei Jiang,
  Bill~Yuchen Lin, Sean Welleck, Peter West, Chandra Bhagavatula, Ronan
  Le~Bras, Jena Hwang, Soumya Sanyal, Xiang Ren, Allyson Ettinger, Zaid
  Harchaoui, and Yejin Choi.
\newblock Faith and fate: Limits of transformers on compositionality.
\newblock In A.~Oh, T.~Naumann, A.~Globerson, K.~Saenko, M.~Hardt, and
  S.~Levine (eds.), \emph{Advances in Neural Information Processing Systems},
  volume~36, pp.\  70293--70332. Curran Associates, Inc., 2023.
\newblock URL
  \url{https://proceedings.neurips.cc/paper_files/paper/2023/file/deb3c28192f979302c157cb653c15e90-Paper-Conference.pdf}.

\bibitem[Feng et~al.(2023)Feng, Zhang, Gu, Ye, He, and Wang]{feng2023revealing}
Guhao Feng, Bohang Zhang, Yuntian Gu, Haotian Ye, Di~He, and Liwei Wang.
\newblock Towards revealing the mystery behind chain of thought: A theoretical
  perspective.
\newblock In \emph{Thirty-seventh Conference on Neural Information Processing
  Systems}, 2023.
\newblock URL \url{https://openreview.net/forum?id=qHrADgAdYu}.

\bibitem[Fodor \& Pylyshyn(1988)Fodor and Pylyshyn]{FODOR19883}
Jerry~A. Fodor and Zenon~W. Pylyshyn.
\newblock Connectionism and cognitive architecture: A critical analysis.
\newblock \emph{Cognition}, 28\penalty0 (1):\penalty0 3--71, 1988.
\newblock ISSN 0010-0277.
\newblock \doi{https://doi.org/10.1016/0010-0277(88)90031-5}.
\newblock URL
  \url{https://www.sciencedirect.com/science/article/pii/0010027788900315}.

\bibitem[Frankland \& Greene(2020)Frankland and
  Greene]{annurev:/content/journals/10.1146/annurev-psych-122216-011829}
Steven~M. Frankland and Joshua~D. Greene.
\newblock Concepts and compositionality: In search of the brain's language of
  thought.
\newblock \emph{Annual Review of Psychology}, 71\penalty0 (Volume 71,
  2020):\penalty0 273--303, 2020.
\newblock ISSN 1545-2085.
\newblock \doi{https://doi.org/10.1146/annurev-psych-122216-011829}.
\newblock URL
  \url{https://www.annualreviews.org/content/journals/10.1146/annurev-psych-122216-011829}.

\bibitem[Furrer et~al.(2021)Furrer, van Zee, Scales, and
  Schärli]{furrer2021compositional}
Daniel Furrer, Marc van Zee, Nathan Scales, and Nathanael Schärli.
\newblock Compositional generalization in semantic parsing: Pre-training vs.
  specialized architectures, 2021.

\bibitem[Gao et~al.(2020)Gao, Huang, and Mooney]{gao2020systematic}
Tong Gao, Qi~Huang, and Raymond Mooney.
\newblock Systematic generalization on g{SCAN} with language conditioned
  embedding.
\newblock In Kam-Fai Wong, Kevin Knight, and Hua Wu (eds.), \emph{Proceedings
  of the 1st Conference of the Asia-Pacific Chapter of the Association for
  Computational Linguistics and the 10th International Joint Conference on
  Natural Language Processing}, pp.\  491--503, Suzhou, China, December 2020.
  Association for Computational Linguistics.
\newblock URL \url{https://aclanthology.org/2020.aacl-main.49}.

\bibitem[Gehring et~al.(2017)Gehring, Auli, Grangier, Yarats, and
  Dauphin]{gehring2017convolutional}
Jonas Gehring, Michael Auli, David Grangier, Denis Yarats, and Yann~N. Dauphin.
\newblock Convolutional sequence to sequence learning.
\newblock In \emph{Proceedings of the 34th International Conference on Machine
  Learning - Volume 70}, ICML'17, pp.\  1243–1252. JMLR.org, 2017.

\bibitem[Geirhos et~al.(2020)Geirhos, Jacobsen, Michaelis, Zemel, Brendel,
  Bethge, and Wichmann]{Geirhos_2020}
Robert Geirhos, Jörn-Henrik Jacobsen, Claudio Michaelis, Richard Zemel,
  Wieland Brendel, Matthias Bethge, and Felix~A. Wichmann.
\newblock Shortcut learning in deep neural networks.
\newblock \emph{Nature Machine Intelligence}, 2\penalty0 (11):\penalty0
  665–673, November 2020.
\newblock ISSN 2522-5839.
\newblock \doi{10.1038/s42256-020-00257-z}.
\newblock URL \url{http://dx.doi.org/10.1038/s42256-020-00257-z}.

\bibitem[Giannakopoulou et~al.(2018)Giannakopoulou, Namjoshi, and
  P{\u{a}}s{\u{a}}reanu]{Giannakopoulou2018}
Dimitra Giannakopoulou, Kedar~S. Namjoshi, and Corina~S. P{\u{a}}s{\u{a}}reanu.
\newblock \emph{Compositional Reasoning}, pp.\  345--383.
\newblock Springer International Publishing, Cham, 2018.
\newblock ISBN 978-3-319-10575-8.
\newblock \doi{10.1007/978-3-319-10575-8_12}.
\newblock URL \url{https://doi.org/10.1007/978-3-319-10575-8_12}.

\bibitem[Girshick et~al.(2011)Girshick, Felzenszwalb, and
  McAllester]{NIPS2011_6faa8040}
Ross Girshick, Pedro Felzenszwalb, and David McAllester.
\newblock Object detection with grammar models.
\newblock In J.~Shawe-Taylor, R.~Zemel, P.~Bartlett, F.~Pereira, and K.Q.
  Weinberger (eds.), \emph{Advances in Neural Information Processing Systems},
  volume~24. Curran Associates, Inc., 2011.
\newblock URL
  \url{https://proceedings.neurips.cc/paper_files/paper/2011/file/6faa8040da20ef399b63a72d0e4ab575-Paper.pdf}.

\bibitem[Gontier et~al.(2020)Gontier, Sinha, Reddy, and
  Pal]{gontier2020measuring}
Nicolas Gontier, Koustuv Sinha, Siva Reddy, and Chris Pal.
\newblock Measuring systematic generalization in neural proof generation with
  transformers.
\newblock In H.~Larochelle, M.~Ranzato, R.~Hadsell, M.F. Balcan, and H.~Lin
  (eds.), \emph{Advances in Neural Information Processing Systems}, volume~33,
  pp.\  22231--22242. Curran Associates, Inc., 2020.
\newblock URL
  \url{https://proceedings.neurips.cc/paper_files/paper/2020/file/fc84ad56f9f547eb89c72b9bac209312-Paper.pdf}.

\bibitem[Gui et~al.(2023)Gui, Chang, Huang, Som, Hauptmann, Gao, and
  Bisk]{gui2023training}
Liangke Gui, Yingshan Chang, Qiuyuan Huang, Subhojit Som, Alexander~G
  Hauptmann, Jianfeng Gao, and Yonatan Bisk.
\newblock Training vision-language transformers from captions.
\newblock \emph{Transactions on Machine Learning Research}, 2023.
\newblock ISSN 2835-8856.
\newblock URL \url{https://openreview.net/forum?id=xLnbSpozWS}.

\bibitem[Gupta \& Kembhavi(2023)Gupta and Kembhavi]{gupta2022visual}
Tanmay Gupta and Aniruddha Kembhavi.
\newblock Visual programming: Compositional visual reasoning without training.
\newblock In \emph{Proceedings of the IEEE/CVF Conference on Computer Vision
  and Pattern Recognition (CVPR)}, pp.\  14953--14962, June 2023.

\bibitem[Hahn(2020)]{Hahn_2020}
Michael Hahn.
\newblock Theoretical limitations of self-attention in neural sequence models.
\newblock \emph{Transactions of the Association for Computational Linguistics},
  8:\penalty0 156–171, December 2020.
\newblock ISSN 2307-387X.
\newblock \doi{10.1162/tacl_a_00306}.
\newblock URL \url{http://dx.doi.org/10.1162/tacl_a_00306}.

\bibitem[Haurilet et~al.(2019)Haurilet, Roitberg, and Stiefelhagen]{8953472}
Monica Haurilet, Alina Roitberg, and Rainer Stiefelhagen.
\newblock It's not about the journey; it's about the destination: Following
  soft paths under question-guidance for visual reasoning.
\newblock In \emph{2019 IEEE/CVF Conference on Computer Vision and Pattern
  Recognition (CVPR)}, pp.\  1930--1939, 2019.
\newblock \doi{10.1109/CVPR.2019.00203}.

\bibitem[He et~al.(2015)He, Zhang, Ren, and Sun]{he2015deep}
Kaiming He, X.~Zhang, Shaoqing Ren, and Jian Sun.
\newblock Deep residual learning for image recognition.
\newblock \emph{2016 IEEE Conference on Computer Vision and Pattern Recognition
  (CVPR)}, pp.\  770--778, 2015.
\newblock URL \url{https://api.semanticscholar.org/CorpusID:206594692}.

\bibitem[He \& McAuley(2016)He and McAuley]{10.1145/2872427.2883037}
Ruining He and Julian McAuley.
\newblock Ups and downs: Modeling the visual evolution of fashion trends with
  one-class collaborative filtering.
\newblock In \emph{Proceedings of the 25th International Conference on World
  Wide Web}, WWW '16, pp.\  507–517, Republic and Canton of Geneva, CHE,
  2016. International World Wide Web Conferences Steering Committee.
\newblock ISBN 9781450341431.
\newblock \doi{10.1145/2872427.2883037}.
\newblock URL \url{https://doi.org/10.1145/2872427.2883037}.

\bibitem[Herzig et~al.(2021)Herzig, Shaw, Chang, Guu, Pasupat, and
  Zhang]{herzig2021unlocking}
Jonathan Herzig, Peter Shaw, Ming-Wei Chang, Kelvin Guu, Panupong Pasupat, and
  Yuan Zhang.
\newblock Unlocking compositional generalization in pre-trained models using
  intermediate representations, 2021.

\bibitem[Hewitt et~al.(2020)Hewitt, Hahn, Ganguli, Liang, and
  Manning]{hewitt-etal-2020-rnns}
John Hewitt, Michael Hahn, Surya Ganguli, Percy Liang, and Christopher~D.
  Manning.
\newblock {RNN}s can generate bounded hierarchical languages with optimal
  memory.
\newblock In \emph{Proceedings of the 2020 Conference on Empirical Methods in
  Natural Language Processing (EMNLP)}, pp.\  1978--2010, Online, November
  2020. Association for Computational Linguistics.
\newblock \doi{10.18653/v1/2020.emnlp-main.156}.
\newblock URL \url{https://aclanthology.org/2020.emnlp-main.156}.

\bibitem[Hiebert(2013)]{book}
J.~Hiebert.
\newblock \emph{Conceptual and procedural knowledge: The case of mathematics}.
\newblock Lawrence Erlbaum Associates, Inc., 08 2013.
\newblock ISBN 9781136559761.
\newblock \doi{10.4324/9780203063538}.

\bibitem[Hochreiter \& Schmidhuber(1997)Hochreiter and
  Schmidhuber]{hochreiter1997long}
Sepp Hochreiter and J{\"u}rgen Schmidhuber.
\newblock Long short-term memory.
\newblock \emph{Neural computation}, 9\penalty0 (8):\penalty0 1735--1780, 1997.

\bibitem[Hofer et~al.(2021)Hofer, Le, Levy, and Tenenbaum]{hofer2021learning}
Matthias Hofer, Tuan~Anh Le, Roger Levy, and Josh Tenenbaum.
\newblock Learning evolved combinatorial symbols with a neuro-symbolic
  generative model, 2021.

\bibitem[Hong et~al.(2021)Hong, Li, Zhu, and Huang]{Hong_2021_ICCV}
Yining Hong, Qing Li, Song-Chun Zhu, and Siyuan Huang.
\newblock Vlgrammar: Grounded grammar induction of vision and language.
\newblock In \emph{Proceedings of the IEEE/CVF International Conference on
  Computer Vision (ICCV)}, pp.\  1665--1674, October 2021.

\bibitem[Hsu et~al.(2023)Hsu, Mao, Tenenbaum, and
  Wu]{hsu2023whatsleftconceptgrounding}
Joy Hsu, Jiayuan Mao, Joshua~B. Tenenbaum, and Jiajun Wu.
\newblock What{\textquoteright}s left? concept grounding with logic-enhanced
  foundation models.
\newblock In \emph{Thirty-seventh Conference on Neural Information Processing
  Systems}, 2023.
\newblock URL \url{https://openreview.net/forum?id=sq4o3tjWaj}.

\bibitem[Huang et~al.(2023)Huang, Yu, Ma, Zhong, Feng, Wang, Chen, Peng, Feng,
  Qin, and Liu]{huang2023surveyhallucinationlargelanguage}
Lei Huang, Weijiang Yu, Weitao Ma, Weihong Zhong, Zhangyin Feng, Haotian Wang,
  Qianglong Chen, Weihua Peng, Xiaocheng Feng, Bing Qin, and Ting Liu.
\newblock A survey on hallucination in large language models: Principles,
  taxonomy, challenges, and open questions, 2023.
\newblock URL \url{https://arxiv.org/abs/2311.05232}.

\bibitem[Hupkes et~al.(2018)Hupkes, Veldhoen, and
  Zuidema]{hupkes2018visualisation}
Dieuwke Hupkes, Sara Veldhoen, and Willem Zuidema.
\newblock Visualisation and ‘diagnostic classifiers’ reveal how recurrent
  and recursive neural networks process hierarchical structure.
\newblock \emph{J. Artif. Int. Res.}, 61\penalty0 (1):\penalty0 907–926,
  January 2018.
\newblock ISSN 1076-9757.

\bibitem[Hupkes et~al.(2020)Hupkes, Dankers, Mul, and
  Bruni]{hupkes2020compositionality}
Dieuwke Hupkes, Verna Dankers, Mathijs Mul, and Elia Bruni.
\newblock Compositionality decomposed: How do neural networks generalise?
  (extended abstract).
\newblock In Christian Bessiere (ed.), \emph{Proceedings of the Twenty-Ninth
  International Joint Conference on Artificial Intelligence, {IJCAI-20}}, pp.\
  5065--5069. International Joint Conferences on Artificial Intelligence
  Organization, 7 2020.
\newblock \doi{10.24963/ijcai.2020/708}.
\newblock URL \url{https://doi.org/10.24963/ijcai.2020/708}.
\newblock Journal track.

\bibitem[Hupkes et~al.(2023)Hupkes, Giulianelli, Dankers, Artetxe, Elazar,
  Pimentel, Christodoulopoulos, Lasri, Saphra, Sinclair, Ulmer, Schottmann,
  Batsuren, Sun, Sinha, Khalatbari, Ryskina, Frieske, Cotterell, and
  Jin]{hupkes2023stateoftheart}
Dieuwke Hupkes, Mario Giulianelli, Verna Dankers, Mikel Artetxe, Yanai Elazar,
  Tiago Pimentel, Christos Christodoulopoulos, Karim Lasri, Naomi Saphra,
  Arabella Sinclair, Dennis Ulmer, Florian Schottmann, Khuyagbaatar Batsuren,
  Kaiser Sun, Koustuv Sinha, Leila Khalatbari, Maria Ryskina, Rita Frieske,
  Ryan Cotterell, and Zhijing Jin.
\newblock A taxonomy and review of generalization research in nlp.
\newblock \emph{Nature Machine Intelligence}, 5\penalty0 (10):\penalty0
  1161--1174, 2023.
\newblock ISSN 2522-5839.
\newblock \doi{10.1038/s42256-023-00729-y}.
\newblock URL \url{https://doi.org/10.1038/s42256-023-00729-y}.

\bibitem[Irsoy \& Cardie(2014)Irsoy and Cardie]{NIPS2014_2cfd4560}
Ozan Irsoy and Claire Cardie.
\newblock Deep recursive neural networks for compositionality in language.
\newblock In Z.~Ghahramani, M.~Welling, C.~Cortes, N.~Lawrence, and K.Q.
  Weinberger (eds.), \emph{Advances in Neural Information Processing Systems},
  volume~27. Curran Associates, Inc., 2014.
\newblock URL
  \url{https://proceedings.neurips.cc/paper_files/paper/2014/file/2cfd4560539f887a5e420412b370b361-Paper.pdf}.

\bibitem[Isola et~al.(2015)Isola, Lim, and Adelson]{StatesAndTransformations}
Phillip Isola, Joseph~J. Lim, and Edward~H. Adelson.
\newblock Discovering states and transformations in image collections.
\newblock In \emph{CVPR}, 2015.

\bibitem[Ito et~al.(2022)Ito, Klinger, Schultz, Murray, Cole, and
  Rigotti]{NEURIPS2022_d0241a0f}
Takuya Ito, Tim Klinger, Doug Schultz, John Murray, Michael Cole, and Mattia
  Rigotti.
\newblock Compositional generalization through abstract representations in
  human and artificial neural networks.
\newblock In S.~Koyejo, S.~Mohamed, A.~Agarwal, D.~Belgrave, K.~Cho, and A.~Oh
  (eds.), \emph{Advances in Neural Information Processing Systems}, volume~35,
  pp.\  32225--32239. Curran Associates, Inc., 2022.
\newblock URL
  \url{https://proceedings.neurips.cc/paper_files/paper/2022/file/d0241a0fb1fc9be477bdfde5e0da276a-Paper-Conference.pdf}.

\bibitem[Janssen \& Partee(1997)Janssen and Partee]{JANSSEN1997417}
Theo~M.V. Janssen and Barbara~H. Partee.
\newblock Chapter 7 - compositionality.
\newblock In Johan {van Benthem} and Alice {ter Meulen} (eds.), \emph{Handbook
  of Logic and Language}, pp.\  417--473. North-Holland, Amsterdam, 1997.
\newblock ISBN 978-0-444-81714-3.
\newblock \doi{https://doi.org/10.1016/B978-044481714-3/50011-4}.
\newblock URL
  \url{https://www.sciencedirect.com/science/article/pii/B9780444817143500114}.

\bibitem[Ji et~al.(2022)Ji, Kojima, Rush, Suhr, Vong, Hawkins, and
  Artzi]{ji2022abstract}
Anya Ji, Noriyuki Kojima, Noah Rush, Alane Suhr, Wai~Keen Vong, Robert Hawkins,
  and Yoav Artzi.
\newblock Abstract visual reasoning with tangram shapes.
\newblock In Yoav Goldberg, Zornitsa Kozareva, and Yue Zhang (eds.),
  \emph{Proceedings of the 2022 Conference on Empirical Methods in Natural
  Language Processing}, pp.\  582--601, Abu Dhabi, United Arab Emirates,
  December 2022. Association for Computational Linguistics.
\newblock \doi{10.18653/v1/2022.emnlp-main.38}.
\newblock URL \url{https://aclanthology.org/2022.emnlp-main.38}.

\bibitem[Jin* et~al.(2023)Jin*, Chen*, Leeb*, Gresele*, Kamal, Lyu, Blin,
  Gonzalez, Kleiman-Weiner, Sachan, and Sch{\"{o}}lkopf]{jin2023cladder}
Zhijing Jin*, Yuen Chen*, Felix Leeb*, Luigi Gresele*, Ojasv Kamal, Zhiheng
  Lyu, Kevin Blin, Fernando Gonzalez, Max Kleiman-Weiner, Mrinmaya Sachan, and
  Bernhard Sch{\"{o}}lkopf.
\newblock Cladder: Assessing causal reasoning in language models.
\newblock In \emph{NeurIPS}, 2023.
\newblock URL \url{https://zhijing-jin.com/files/papers/CLadder_2023.pdf}.

\bibitem[Johnson et~al.(2016)Johnson, Hariharan, van~der Maaten, Fei-Fei,
  Zitnick, and Girshick]{johnson2016clevr}
Justin Johnson, Bharath Hariharan, Laurens van~der Maaten, Li~Fei-Fei,
  C.~Lawrence Zitnick, and Ross~B. Girshick.
\newblock Clevr: A diagnostic dataset for compositional language and elementary
  visual reasoning.
\newblock \emph{2017 IEEE Conference on Computer Vision and Pattern Recognition
  (CVPR)}, pp.\  1988--1997, 2016.
\newblock URL \url{https://api.semanticscholar.org/CorpusID:15458100}.

\bibitem[Kamali \& Kordjamshidi(2023)Kamali and
  Kordjamshidi]{kamali2023syntaxguided}
Danial Kamali and Parisa Kordjamshidi.
\newblock Syntax-guided transformers: Elevating compositional generalization
  and grounding in multimodal environments.
\newblock In Dieuwke Hupkes, Verna Dankers, Khuyagbaatar Batsuren, Koustuv
  Sinha, Amirhossein Kazemnejad, Christos Christodoulopoulos, Ryan Cotterell,
  and Elia Bruni (eds.), \emph{Proceedings of the 1st GenBench Workshop on
  (Benchmarking) Generalisation in NLP}, pp.\  130--142, Singapore, December
  2023. Association for Computational Linguistics.
\newblock \doi{10.18653/v1/2023.genbench-1.10}.
\newblock URL \url{https://aclanthology.org/2023.genbench-1.10}.

\bibitem[Kazemnejad et~al.(2023)Kazemnejad, Padhi, Natesan, Das, and
  Reddy]{kazemnejad2023impactpositionalencodinglength}
Amirhossein Kazemnejad, Inkit Padhi, Karthikeyan Natesan, Payel Das, and Siva
  Reddy.
\newblock The impact of positional encoding on length generalization in
  transformers.
\newblock In \emph{Thirty-seventh Conference on Neural Information Processing
  Systems}, 2023.
\newblock URL \url{https://openreview.net/forum?id=Drrl2gcjzl}.

\bibitem[Keysers et~al.(2020)Keysers, Sch{\"a}rli, Scales, Buisman, Furrer,
  Kashubin, Momchev, Sinopalnikov, Stafiniak, Tihon, Tsarkov, Wang, van Zee,
  and Bousquet]{keysers2020measuring}
Daniel Keysers, Nathanael Sch{\"a}rli, Nathan Scales, Hylke Buisman, Daniel
  Furrer, Sergii Kashubin, Nikola Momchev, Danila Sinopalnikov, Lukasz
  Stafiniak, Tibor Tihon, Dmitry Tsarkov, Xiao Wang, Marc van Zee, and Olivier
  Bousquet.
\newblock Measuring compositional generalization: A comprehensive method on
  realistic data.
\newblock In \emph{International Conference on Learning Representations}, 2020.
\newblock URL \url{https://openreview.net/forum?id=SygcCnNKwr}.

\bibitem[Khot et~al.(2023)Khot, Trivedi, Finlayson, Fu, Richardson, Clark, and
  Sabharwal]{khot2023decomposed}
Tushar Khot, Harsh Trivedi, Matthew Finlayson, Yao Fu, Kyle Richardson, Peter
  Clark, and Ashish Sabharwal.
\newblock Decomposed prompting: A modular approach for solving complex tasks.
\newblock In \emph{The Eleventh International Conference on Learning
  Representations}, 2023.
\newblock URL \url{https://openreview.net/forum?id=_nGgzQjzaRy}.

\bibitem[Kim \& Linzen(2020)Kim and Linzen]{kim2020cogs}
Najoung Kim and Tal Linzen.
\newblock {COGS}: A compositional generalization challenge based on semantic
  interpretation.
\newblock In Bonnie Webber, Trevor Cohn, Yulan He, and Yang Liu (eds.),
  \emph{Proceedings of the 2020 Conference on Empirical Methods in Natural
  Language Processing (EMNLP)}, pp.\  9087--9105, Online, November 2020.
  Association for Computational Linguistics.
\newblock \doi{10.18653/v1/2020.emnlp-main.731}.
\newblock URL \url{https://aclanthology.org/2020.emnlp-main.731}.

\bibitem[Kleinberg \& Tardos(2005)Kleinberg and Tardos]{10.5555/1051910}
Jon Kleinberg and Eva Tardos.
\newblock \emph{Algorithm Design}.
\newblock Addison-Wesley Longman Publishing Co., Inc., USA, 2005.
\newblock ISBN 0321295358.

\bibitem[Klinger et~al.(2020)Klinger, Adjodah, Marois, Joseph, Riemer,
  Pentland, and Campbell]{klinger2020study}
Tim Klinger, Dhaval Adjodah, Vincent Marois, Josh Joseph, Matthew Riemer,
  Alex~'Sandy' Pentland, and Murray Campbell.
\newblock A study of compositional generalization in neural models, 2020.

\bibitem[Klinger et~al.(2024)Klinger, Liu, Dan, Crouse, Ram, and
  Gray]{klinger2024compositional}
Tim Klinger, Luke Liu, Soham Dan, Maxwell Crouse, Parikshit Ram, and Alexander
  Gray.
\newblock Compositional program generation for few-shot systematic
  generalization, 2024.

\bibitem[Kojima et~al.(2023)Kojima, Averbuch-Elor, and Artzi]{kojima2023joint}
Noriyuki Kojima, Hadar Averbuch-Elor, and Yoav Artzi.
\newblock A joint study of phrase grounding and task performance in vision and
  language models.
\newblock \emph{Trans. Mach. Learn. Res.}, 2024, 2023.
\newblock URL \url{https://api.semanticscholar.org/CorpusID:261556878}.

\bibitem[Korrel et~al.(2019)Korrel, Hupkes, Dankers, and
  Bruni]{korrel2019transcoding}
Kris Korrel, Dieuwke Hupkes, Verna Dankers, and Elia Bruni.
\newblock Transcoding compositionally: Using attention to find more
  generalizable solutions.
\newblock In Tal Linzen, Grzegorz Chrupa{\l}a, Yonatan Belinkov, and Dieuwke
  Hupkes (eds.), \emph{Proceedings of the 2019 ACL Workshop BlackboxNLP:
  Analyzing and Interpreting Neural Networks for NLP}, pp.\  1--11, Florence,
  Italy, August 2019. Association for Computational Linguistics.
\newblock \doi{10.18653/v1/W19-4801}.
\newblock URL \url{https://aclanthology.org/W19-4801}.

\bibitem[Kuo et~al.(2020)Kuo, Katz, and
  Barbu]{https://doi.org/10.48550/arxiv.2008.02742}
Yen-Ling Kuo, Boris Katz, and Andrei Barbu.
\newblock Compositional networks enable systematic generalization for grounded
  language understanding, 2020.
\newblock URL \url{https://arxiv.org/abs/2008.02742}.

\bibitem[Lake \& Baroni(2017)Lake and Baroni]{lake2018generalization}
Brenden~M. Lake and Marco Baroni.
\newblock Generalization without systematicity: On the compositional skills of
  sequence-to-sequence recurrent networks.
\newblock In \emph{International Conference on Machine Learning}, 2017.
\newblock URL \url{https://api.semanticscholar.org/CorpusID:46761158}.

\bibitem[Lake et~al.(2017)Lake, Ullman, Tenenbaum, and
  Gershman]{Lake2017-LAKBMT}
Brenden~M. Lake, Tomer~D. Ullman, Joshua~B. Tenenbaum, and Samuel~J. Gershman.
\newblock Building machines that learn and think like people.
\newblock \emph{Behavioral and Brain Sciences}, 40, 2017.
\newblock \doi{10.1017/s0140525x16001837}.

\bibitem[Lake et~al.(2019)Lake, Linzen, and Baroni]{lake2019human}
Brenden~M. Lake, Tal Linzen, and Marco Baroni.
\newblock Human few-shot learning of compositional instructions.
\newblock In \emph{Annual Meeting of the Cognitive Science Society}, 2019.
\newblock URL \url{https://api.semanticscholar.org/CorpusID:58006558}.

\bibitem[Lample et~al.(2019)Lample, Subramanian, Smith, Denoyer, Ranzato, and
  Boureau]{lample2018multipleattribute}
Guillaume Lample, Sandeep Subramanian, Eric Smith, Ludovic Denoyer,
  Marc'Aurelio Ranzato, and Y-Lan Boureau.
\newblock Multiple-attribute text rewriting.
\newblock In \emph{International Conference on Learning Representations}, 2019.
\newblock URL \url{https://openreview.net/forum?id=H1g2NhC5KQ}.

\bibitem[LeCun et~al.(1989)LeCun, Boser, Denker, Henderson, Howard, Hubbard,
  and Jackel]{NIPS1989_53c3bce6}
Yann LeCun, Bernhard Boser, John Denker, Donnie Henderson, R.~Howard, Wayne
  Hubbard, and Lawrence Jackel.
\newblock Handwritten digit recognition with a back-propagation network.
\newblock In D.~Touretzky (ed.), \emph{Advances in Neural Information
  Processing Systems}, volume~2. Morgan-Kaufmann, 1989.
\newblock URL
  \url{https://proceedings.neurips.cc/paper_files/paper/1989/file/53c3bce66e43be4f209556518c2fcb54-Paper.pdf}.

\bibitem[Lepori et~al.(2023)Lepori, Serre, and Pavlick]{lepori2023break}
Michael~A. Lepori, Thomas Serre, and Ellie Pavlick.
\newblock Break it down: Evidence for structural compositionality in neural
  networks.
\newblock In \emph{Thirty-seventh Conference on Neural Information Processing
  Systems}, 2023.
\newblock URL \url{https://openreview.net/forum?id=rwbzMiuFQl}.

\bibitem[Li et~al.(2021)Li, Yin, Chen, and Zhang]{li-etal-2021-compositional}
Yafu Li, Yongjing Yin, Yulong Chen, and Yue Zhang.
\newblock On compositional generalization of neural machine translation.
\newblock In Chengqing Zong, Fei Xia, Wenjie Li, and Roberto Navigli (eds.),
  \emph{Proceedings of the 59th Annual Meeting of the Association for
  Computational Linguistics and the 11th International Joint Conference on
  Natural Language Processing (Volume 1: Long Papers)}, pp.\  4767--4780,
  Online, August 2021. Association for Computational Linguistics.
\newblock \doi{10.18653/v1/2021.acl-long.368}.
\newblock URL \url{https://aclanthology.org/2021.acl-long.368}.

\bibitem[Liao et~al.(2023)Liao, Wei, Jiang, Zhang, and
  Ishibuchi]{NEURIPS2023_6a42b45a}
Weiduo Liao, Ying Wei, Mingchen Jiang, Qingfu Zhang, and Hisao Ishibuchi.
\newblock Does continual learning meet compositionality? new benchmarks and an
  evaluation framework.
\newblock In A.~Oh, T.~Naumann, A.~Globerson, K.~Saenko, M.~Hardt, and
  S.~Levine (eds.), \emph{Advances in Neural Information Processing Systems},
  volume~36, pp.\  33499--33513. Curran Associates, Inc., 2023.
\newblock URL
  \url{https://proceedings.neurips.cc/paper_files/paper/2023/file/6a42b45af2b72e6e5b5e3a6fe695809f-Paper-Datasets_and_Benchmarks.pdf}.

\bibitem[Lin et~al.(2024)Lin, Chen, Pathak, Zhang, and
  Ramanan]{lin2023revisiting}
Zhiqiu Lin, Xinyue Chen, Deepak Pathak, Pengchuan Zhang, and Deva Ramanan.
\newblock Revisiting the role of language priors in vision-language models.
\newblock In \emph{ICML}, 2024.
\newblock URL \url{https://openreview.net/forum?id=J5VB1h3Aed}.

\bibitem[Ling et~al.(2017)Ling, Yogatama, Dyer, and Blunsom]{ling2017program}
Wang Ling, Dani Yogatama, Chris Dyer, and Phil Blunsom.
\newblock Program induction by rationale generation: Learning to solve and
  explain algebraic word problems.
\newblock In Regina Barzilay and Min-Yen Kan (eds.), \emph{Proceedings of the
  55th Annual Meeting of the Association for Computational Linguistics (Volume
  1: Long Papers)}, pp.\  158--167, Vancouver, Canada, July 2017. Association
  for Computational Linguistics.
\newblock \doi{10.18653/v1/P17-1015}.
\newblock URL \url{https://aclanthology.org/P17-1015}.

\bibitem[Liu et~al.(2021)Liu, An, Lin, Liu, Chen, Lou, Wen, Zheng, and
  Zhang]{liu2021learning}
Chenyao Liu, Shengnan An, Zeqi Lin, Qian Liu, Bei Chen, Jian-Guang Lou, Lijie
  Wen, Nanning Zheng, and Dongmei Zhang.
\newblock Learning algebraic recombination for compositional generalization.
\newblock In Chengqing Zong, Fei Xia, Wenjie Li, and Roberto Navigli (eds.),
  \emph{Findings of the Association for Computational Linguistics: ACL-IJCNLP
  2021}, pp.\  1129--1144, Online, August 2021. Association for Computational
  Linguistics.
\newblock \doi{10.18653/v1/2021.findings-acl.97}.
\newblock URL \url{https://aclanthology.org/2021.findings-acl.97}.

\bibitem[Liu et~al.(2022{\natexlab{a}})Liu, Li, Guo, Luo, and
  Wang]{liu-etal-2022-multi-attribute}
Guisheng Liu, Yi~Li, Yanqing Guo, Xiangyang Luo, and Bo~Wang.
\newblock Multi-attribute controlled text generation with contrastive-generator
  and external-discriminator.
\newblock In Nicoletta Calzolari, Chu-Ren Huang, Hansaem Kim, James
  Pustejovsky, Leo Wanner, Key-Sun Choi, Pum-Mo Ryu, Hsin-Hsi Chen, Lucia
  Donatelli, Heng Ji, Sadao Kurohashi, Patrizia Paggio, Nianwen Xue, Seokhwan
  Kim, Younggyun Hahm, Zhong He, Tony~Kyungil Lee, Enrico Santus, Francis Bond,
  and Seung-Hoon Na (eds.), \emph{Proceedings of the 29th International
  Conference on Computational Linguistics}, pp.\  5904--5913, Gyeongju,
  Republic of Korea, October 2022{\natexlab{a}}. International Committee on
  Computational Linguistics.
\newblock URL \url{https://aclanthology.org/2022.coling-1.516}.

\bibitem[Liu et~al.(2022{\natexlab{b}})Liu, Li, Du, Torralba, and
  Tenenbaum]{liu2023compositional}
Nan Liu, Shuang Li, Yilun Du, Antonio Torralba, and Joshua~B. Tenenbaum.
\newblock Compositional visual generation with composable diffusion models.
\newblock In \emph{Computer Vision – ECCV 2022: 17th European Conference, Tel
  Aviv, Israel, October 23–27, 2022, Proceedings, Part XVII}, pp.\
  423–439, Berlin, Heidelberg, 2022{\natexlab{b}}. Springer-Verlag.
\newblock ISBN 978-3-031-19789-5.
\newblock \doi{10.1007/978-3-031-19790-1_26}.
\newblock URL \url{https://doi.org/10.1007/978-3-031-19790-1_26}.

\bibitem[Lu et~al.(2024)Lu, Bigoulaeva, Sachdeva, Tayyar~Madabushi, and
  Gurevych]{lu2023emergent}
Sheng Lu, Irina Bigoulaeva, Rachneet Sachdeva, Harish Tayyar~Madabushi, and
  Iryna Gurevych.
\newblock Are emergent abilities in large language models just in-context
  learning?
\newblock In Lun-Wei Ku, Andre Martins, and Vivek Srikumar (eds.),
  \emph{Proceedings of the 62nd Annual Meeting of the Association for
  Computational Linguistics (Volume 1: Long Papers)}, pp.\  5098--5139,
  Bangkok, Thailand, August 2024. Association for Computational Linguistics.
\newblock \doi{10.18653/v1/2024.acl-long.279}.
\newblock URL \url{https://aclanthology.org/2024.acl-long.279}.

\bibitem[Ma et~al.(2022)Ma, Hong, Gul, Gandhi, Gao, and Krishna]{ma2023crepe}
Zixian Ma, Jerry Hong, Mustafa~Omer Gul, Mona Gandhi, Irena Gao, and Ranjay
  Krishna.
\newblock @ crepe: Can vision-language foundation models reason
  compositionally?
\newblock \emph{2023 IEEE/CVF Conference on Computer Vision and Pattern
  Recognition (CVPR)}, pp.\  10910--10921, 2022.
\newblock URL \url{https://api.semanticscholar.org/CorpusID:254685851}.

\bibitem[Marcus(2018)]{marcus2018deep}
Gary Marcus.
\newblock Deep learning: A critical appraisal, 2018.

\bibitem[Marcus et~al.(1992)Marcus, Pinker, Ullman, Hollander, Rosen, Xu, and
  Clahsen]{eadfa2ba-d727-3995-86be-56ba36e587c4}
Gary~F. Marcus, Steven Pinker, Michael Ullman, Michelle Hollander, T.~John
  Rosen, Fei Xu, and Harald Clahsen.
\newblock Overregularization in language acquisition.
\newblock \emph{Monographs of the Society for Research in Child Development},
  57\penalty0 (4):\penalty0 i--178, 1992.
\newblock ISSN 0037976X, 15405834.
\newblock URL \url{http://www.jstor.org/stable/1166115}.

\bibitem[Marelli et~al.(2014)Marelli, Menini, Baroni, Bentivogli, Bernardi, and
  Zamparelli]{marelli-etal-2014-sick}
Marco Marelli, Stefano Menini, Marco Baroni, Luisa Bentivogli, Raffaella
  Bernardi, and Roberto Zamparelli.
\newblock A {SICK} cure for the evaluation of compositional distributional
  semantic models.
\newblock In Nicoletta Calzolari, Khalid Choukri, Thierry Declerck, Hrafn
  Loftsson, Bente Maegaard, Joseph Mariani, Asuncion Moreno, Jan Odijk, and
  Stelios Piperidis (eds.), \emph{Proceedings of the Ninth International
  Conference on Language Resources and Evaluation ({LREC}'14)}, pp.\  216--223,
  Reykjavik, Iceland, May 2014. European Language Resources Association (ELRA).
\newblock URL
  \url{http://www.lrec-conf.org/proceedings/lrec2014/pdf/363_Paper.pdf}.

\bibitem[Mendez \& Eaton(2021)Mendez and
  Eaton]{mendez2021lifelonglearningcompositionalstructures}
Jorge~A Mendez and Eric Eaton.
\newblock Lifelong learning of compositional structures.
\newblock In \emph{International Conference on Learning Representations}, 2021.
\newblock URL \url{https://openreview.net/forum?id=ADWd4TJO13G}.

\bibitem[Mendez \& Eaton(2023)Mendez and
  Eaton]{mendez2023reusecomposeknowledgelifetime}
Jorge~A. Mendez and Eric Eaton.
\newblock How to reuse and compose knowledge for a lifetime of tasks: A survey
  on continual learning and functional composition, 2023.
\newblock URL \url{https://arxiv.org/abs/2207.07730}.

\bibitem[Merrill \& Sabharwal(2023)Merrill and
  Sabharwal]{merrill2023parallelism}
William Merrill and Ashish Sabharwal.
\newblock The parallelism tradeoff: Limitations of log-precision transformers.
\newblock \emph{Transactions of the Association for Computational Linguistics},
  11:\penalty0 531--545, 2023.
\newblock \doi{10.1162/tacl_a_00562}.
\newblock URL \url{https://aclanthology.org/2023.tacl-1.31}.

\bibitem[Mikolov et~al.(2018)Mikolov, Joulin, and Baroni]{mikolov2016roadmap}
Tomas Mikolov, Armand Joulin, and Marco Baroni.
\newblock A roadmap towards machine intelligence.
\newblock In Alexander Gelbukh (ed.), \emph{Computational Linguistics and
  Intelligent Text Processing}, pp.\  29--61, Cham, 2018. Springer
  International Publishing.
\newblock ISBN 978-3-319-75477-2.

\bibitem[Minervini et~al.(2020)Minervini, Riedel, Stenetorp, Grefenstette, and
  Rockt\"{a}schel]{minervini2020learning}
Pasquale Minervini, Sebastian Riedel, Pontus Stenetorp, Edward Grefenstette,
  and Tim Rockt\"{a}schel.
\newblock Learning reasoning strategies in end-to-end differentiable proving.
\newblock In \emph{Proceedings of the 37th International Conference on Machine
  Learning}, ICML'20. JMLR.org, 2020.

\bibitem[Mirzaee \& Kordjamshidi(2022)Mirzaee and
  Kordjamshidi]{mirzaee-kordjamshidi-2022-transfer}
Roshanak Mirzaee and Parisa Kordjamshidi.
\newblock Transfer learning with synthetic corpora for spatial role labeling
  and reasoning.
\newblock In Yoav Goldberg, Zornitsa Kozareva, and Yue Zhang (eds.),
  \emph{Proceedings of the 2022 Conference on Empirical Methods in Natural
  Language Processing}, pp.\  6148--6165, Abu Dhabi, United Arab Emirates,
  December 2022. Association for Computational Linguistics.
\newblock \doi{10.18653/v1/2022.emnlp-main.413}.
\newblock URL \url{https://aclanthology.org/2022.emnlp-main.413}.

\bibitem[Mirzaee et~al.(2021)Mirzaee, Rajaby~Faghihi, Ning, and
  Kordjamshidi]{mirzaee-etal-2021-spartqa}
Roshanak Mirzaee, Hossein Rajaby~Faghihi, Qiang Ning, and Parisa Kordjamshidi.
\newblock {SPARTQA}: A textual question answering benchmark for spatial
  reasoning.
\newblock In Kristina Toutanova, Anna Rumshisky, Luke Zettlemoyer, Dilek
  Hakkani-Tur, Iz~Beltagy, Steven Bethard, Ryan Cotterell, Tanmoy Chakraborty,
  and Yichao Zhou (eds.), \emph{Proceedings of the 2021 Conference of the North
  American Chapter of the Association for Computational Linguistics: Human
  Language Technologies}, pp.\  4582--4598, Online, June 2021. Association for
  Computational Linguistics.
\newblock \doi{10.18653/v1/2021.naacl-main.364}.
\newblock URL \url{https://aclanthology.org/2021.naacl-main.364}.

\bibitem[Montague(1974)]{Montague1974-MONFPS}
Richard Montague.
\newblock \emph{Formal Philosophy: Selected Papers of Richard Montague}.
\newblock Yale University Press, New Haven,, 1974.

\bibitem[Murty et~al.(2023)Murty, Sharma, Andreas, and
  Manning]{murty2023pushdown}
Shikhar Murty, Pratyusha Sharma, Jacob Andreas, and Christopher Manning.
\newblock Pushdown layers: Encoding recursive structure in transformer language
  models.
\newblock In Houda Bouamor, Juan Pino, and Kalika Bali (eds.),
  \emph{Proceedings of the 2023 Conference on Empirical Methods in Natural
  Language Processing}, pp.\  3233--3247, Singapore, December 2023. Association
  for Computational Linguistics.
\newblock \doi{10.18653/v1/2023.emnlp-main.195}.
\newblock URL \url{https://aclanthology.org/2023.emnlp-main.195}.

\bibitem[Naeem et~al.(2021)Naeem, Xian, Tombari, and
  Akata]{naeem2021learninggraphembeddingscompositional}
Muhammad~Ferjad Naeem, Yongqin Xian, Federico Tombari, and Zeynep Akata.
\newblock Learning graph embeddings for compositional zero-shot learning.
\newblock \emph{2021 IEEE/CVF Conference on Computer Vision and Pattern
  Recognition (CVPR)}, pp.\  953--962, 2021.
\newblock URL \url{https://api.semanticscholar.org/CorpusID:231786709}.

\bibitem[Nafar et~al.(2024{\natexlab{a}})Nafar, Venable, and
  Kordjamshidi]{nafar-etal-2024-teaching}
Aliakbar Nafar, K.~Brent Venable, and Parisa Kordjamshidi.
\newblock Teaching probabilistic logical reasoning to transformers.
\newblock In Yvette Graham and Matthew Purver (eds.), \emph{Findings of the
  Association for Computational Linguistics: EACL 2024}, pp.\  1615--1632, St.
  Julian{'}s, Malta, March 2024{\natexlab{a}}. Association for Computational
  Linguistics.
\newblock URL \url{https://aclanthology.org/2024.findings-eacl.112}.

\bibitem[Nafar et~al.(2024{\natexlab{b}})Nafar, Venable, and
  Kordjamshidi]{nafar2024probabilistic}
Aliakbar Nafar, Kristen~Brent Venable, and Parisa Kordjamshidi.
\newblock Probabilistic reasoning in generative large language models,
  2024{\natexlab{b}}.
\newblock URL \url{https://arxiv.org/abs/2402.09614}.

\bibitem[Niemeyer \& Geiger(2021)Niemeyer and Geiger]{niemeyer2021giraffe}
Michael Niemeyer and Andreas Geiger.
\newblock { GIRAFFE: Representing Scenes as Compositional Generative Neural
  Feature Fields }.
\newblock In \emph{2021 IEEE/CVF Conference on Computer Vision and Pattern
  Recognition (CVPR)}, pp.\  11448--11459, Los Alamitos, CA, USA, June 2021.
  IEEE Computer Society.
\newblock \doi{10.1109/CVPR46437.2021.01129}.
\newblock URL
  \url{https://doi.ieeecomputersociety.org/10.1109/CVPR46437.2021.01129}.

\bibitem[Nye et~al.(2020)Nye, Solar-Lezama, Tenenbaum, and
  Lake]{nye2020learning}
Maxwell~I. Nye, Armando Solar-Lezama, Joshua~B. Tenenbaum, and Brenden~M. Lake.
\newblock Learning compositional rules via neural program synthesis.
\newblock In \emph{Proceedings of the 34th International Conference on Neural
  Information Processing Systems}, NIPS '20, Red Hook, NY, USA, 2020. Curran
  Associates Inc.
\newblock ISBN 9781713829546.

\bibitem[Okawa et~al.(2023)Okawa, Lubana, Dick, and
  Tanaka]{okawa2023compositional}
Maya Okawa, Ekdeep~Singh Lubana, Robert~P. Dick, and Hidenori Tanaka.
\newblock Compositional abilities emerge multiplicatively: Exploring diffusion
  models on a synthetic task.
\newblock In \emph{Thirty-seventh Conference on Neural Information Processing
  Systems}, 2023.
\newblock URL \url{https://openreview.net/forum?id=frVo9MzRuU}.

\bibitem[Ontanon et~al.(2022)Ontanon, Ainslie, Fisher, and
  Cvicek]{https://doi.org/10.48550/arxiv.2108.04378}
Santiago Ontanon, Joshua Ainslie, Zachary Fisher, and Vaclav Cvicek.
\newblock Making transformers solve compositional tasks.
\newblock In Smaranda Muresan, Preslav Nakov, and Aline Villavicencio (eds.),
  \emph{Proceedings of the 60th Annual Meeting of the Association for
  Computational Linguistics (Volume 1: Long Papers)}, pp.\  3591--3607, Dublin,
  Ireland, May 2022. Association for Computational Linguistics.
\newblock \doi{10.18653/v1/2022.acl-long.251}.
\newblock URL \url{https://aclanthology.org/2022.acl-long.251}.

\bibitem[OpenAI(2024)]{openai2024gpt4}
OpenAI.
\newblock {GPT-4} {T}echnical {R}eport, 2024.

\bibitem[Ouyang et~al.(2024)Ouyang, Wu, Jiang, Almeida, Wainwright, Mishkin,
  Zhang, Agarwal, Slama, Ray, Schulman, Hilton, Kelton, Miller, Simens, Askell,
  Welinder, Christiano, Leike, and
  Lowe]{ouyang2022traininglanguagemodelsfollow}
Long Ouyang, Jeff Wu, Xu~Jiang, Diogo Almeida, Carroll~L. Wainwright, Pamela
  Mishkin, Chong Zhang, Sandhini Agarwal, Katarina Slama, Alex Ray, John
  Schulman, Jacob Hilton, Fraser Kelton, Luke Miller, Maddie Simens, Amanda
  Askell, Peter Welinder, Paul Christiano, Jan Leike, and Ryan Lowe.
\newblock Training language models to follow instructions with human feedback.
\newblock In \emph{Proceedings of the 36th International Conference on Neural
  Information Processing Systems}, NIPS '22, Red Hook, NY, USA, 2024. Curran
  Associates Inc.
\newblock ISBN 9781713871088.

\bibitem[Partee(2004)]{Partee2004-PARCIF}
Barbara~Hall Partee.
\newblock \emph{Compositionality in Formal Semantics: Selected Papers of
  Barbara H. Partee}.
\newblock Blackwell, Malden, MA, 2004.

\bibitem[Pearl(1988)]{pearlBook}
Judea Pearl.
\newblock \emph{Probabilistic Reasoning in Intelligent Systems}.
\newblock Morgan-Kaufmann, 1988.

\bibitem[Poesia et~al.(2023)Poesia, Gandhi, Zelikman, and
  Goodman]{poesia2023certified}
Gabriel Poesia, Kanishk Gandhi, Eric Zelikman, and Noah~D. Goodman.
\newblock Certified reasoning with language models, 2023.

\bibitem[Porto(2002)]{10.1007/3-540-45788-7_17}
Ant{\'o}nio Porto.
\newblock Structural abstraction and application in logic programming.
\newblock In Zhenjiang Hu and Mario Rodr{\'i}guez-Artalejo (eds.),
  \emph{Functional and Logic Programming}, pp.\  275--289, Berlin, Heidelberg,
  2002. Springer Berlin Heidelberg.
\newblock ISBN 978-3-540-45788-6.

\bibitem[Premsri \& Kordjamshidi(2024)Premsri and
  Kordjamshidi]{premsri2024neurosymbolictrainingreasoningspatial}
Tanawan Premsri and Parisa Kordjamshidi.
\newblock Neuro-symbolic training for reasoning over spatial language, 2024.
\newblock URL \url{https://arxiv.org/abs/2406.13828}.

\bibitem[Press et~al.(2023)Press, Zhang, Min, Schmidt, Smith, and
  Lewis]{press2023measuring}
Ofir Press, Muru Zhang, Sewon Min, Ludwig Schmidt, Noah~A. Smith, and Mike
  Lewis.
\newblock Measuring and narrowing the compositionality gap in language models.
\newblock In \emph{The 2023 Conference on Empirical Methods in Natural Language
  Processing}, 2023.
\newblock URL \url{https://openreview.net/forum?id=feiAVaSXdb}.

\bibitem[Prosser(1993)]{https://doi.org/10.1111/j.1467-8640.1993.tb00310.x}
Patrick Prosser.
\newblock Hybrid algorithms for the constraint satisfaction problem.
\newblock \emph{Computational Intelligence}, 9\penalty0 (3):\penalty0 268--299,
  1993.
\newblock \doi{https://doi.org/10.1111/j.1467-8640.1993.tb00310.x}.
\newblock URL
  \url{https://onlinelibrary.wiley.com/doi/abs/10.1111/j.1467-8640.1993.tb00310.x}.

\bibitem[Pérez et~al.(2019)Pérez, Marinković, and Barceló]{perez2018on}
Jorge Pérez, Javier Marinković, and Pablo Barceló.
\newblock On the turing completeness of modern neural network architectures.
\newblock In \emph{International Conference on Learning Representations}, 2019.
\newblock URL \url{https://openreview.net/forum?id=HyGBdo0qFm}.

\bibitem[Pérez et~al.(2021)Pérez, Barceló, and Marinkovic]{JMLR:v22:20-302}
Jorge Pérez, Pablo Barceló, and Javier Marinkovic.
\newblock Attention is turing-complete.
\newblock \emph{Journal of Machine Learning Research}, 22\penalty0
  (75):\penalty0 1--35, 2021.
\newblock URL \url{http://jmlr.org/papers/v22/20-302.html}.

\bibitem[Qiao et~al.(2023)Qiao, Ou, Zhang, Chen, Yao, Deng, Tan, Huang, and
  Chen]{qiao2023reasoning}
Shuofei Qiao, Yixin Ou, Ningyu Zhang, Xiang Chen, Yunzhi Yao, Shumin Deng,
  Chuanqi Tan, Fei Huang, and Huajun Chen.
\newblock Reasoning with language model prompting: A survey.
\newblock In Anna Rogers, Jordan Boyd-Graber, and Naoaki Okazaki (eds.),
  \emph{Proceedings of the 61st Annual Meeting of the Association for
  Computational Linguistics (Volume 1: Long Papers)}, pp.\  5368--5393,
  Toronto, Canada, July 2023. Association for Computational Linguistics.
\newblock \doi{10.18653/v1/2023.acl-long.294}.
\newblock URL \url{https://aclanthology.org/2023.acl-long.294}.

\bibitem[Qiu et~al.(2021)Qiu, Hu, Zhang, Shaw, and
  Sha]{qiu-etal-2021-systematic}
Linlu Qiu, Hexiang Hu, Bowen Zhang, Peter Shaw, and Fei Sha.
\newblock Systematic generalization on g{SCAN}: {W}hat is nearly solved and
  what is next?
\newblock In \emph{Proceedings of the 2021 Conference on Empirical Methods in
  Natural Language Processing}, pp.\  2180--2188, Online and Punta Cana,
  Dominican Republic, November 2021. Association for Computational Linguistics.
\newblock \doi{10.18653/v1/2021.emnlp-main.166}.
\newblock URL \url{https://aclanthology.org/2021.emnlp-main.166}.

\bibitem[Rajaby~Faghihi et~al.(2021)Rajaby~Faghihi, Guo, Uszok, Nafar, and
  Kordjamshidi]{rajaby-faghihi-etal-2021-domiknows}
Hossein Rajaby~Faghihi, Quan Guo, Andrzej Uszok, Aliakbar Nafar, and Parisa
  Kordjamshidi.
\newblock {D}omi{K}now{S}: A library for integration of symbolic domain
  knowledge in deep learning.
\newblock In Heike Adel and Shuming Shi (eds.), \emph{Proceedings of the 2021
  Conference on Empirical Methods in Natural Language Processing: System
  Demonstrations}, pp.\  231--241, Online and Punta Cana, Dominican Republic,
  November 2021. Association for Computational Linguistics.
\newblock \doi{10.18653/v1/2021.emnlp-demo.27}.
\newblock URL \url{https://aclanthology.org/2021.emnlp-demo.27}.

\bibitem[Ram et~al.(2023)Ram, Klinger, and Gray]{ram2023how}
Parikshit Ram, Tim Klinger, and Alexander~G. Gray.
\newblock How compositional is a model?
\newblock In \emph{International Joint Conference on Artificial Intelligence
  2023 Workshop on Knowledge-Based Compositional Generalization}, 2023.
\newblock URL \url{https://openreview.net/forum?id=OImyRhNLv3}.

\bibitem[Ram et~al.(2024)Ram, Klinger, and Gray]{ram2024makes}
Parikshit Ram, Tim Klinger, and Alexander~G. Gray.
\newblock What makes models compositional? a theoretical view.
\newblock In Kate Larson (ed.), \emph{Proceedings of the Thirty-Third
  International Joint Conference on Artificial Intelligence, {IJCAI-24}}, pp.\
  4824--4832. International Joint Conferences on Artificial Intelligence
  Organization, 8 2024.
\newblock \doi{10.24963/ijcai.2024/533}.
\newblock URL \url{https://doi.org/10.24963/ijcai.2024/533}.
\newblock Main Track.

\bibitem[Ruis et~al.(2020)Ruis, Andreas, Baroni, Bouchacourt, and
  Lake]{https://doi.org/10.48550/arxiv.2003.05161}
Laura Ruis, Jacob Andreas, Marco Baroni, Diane Bouchacourt, and Brenden~M.
  Lake.
\newblock A benchmark for systematic generalization in grounded language
  understanding, 2020.
\newblock URL \url{https://arxiv.org/abs/2003.05161}.

\bibitem[Ruoss et~al.(2023)Ruoss, Del{\'e}tang, Genewein, Grau-Moya,
  Csord{\'a}s, Bennani, Legg, and
  Veness]{ruoss2023randomizedpositionalencodingsboost}
Anian Ruoss, Gr{\'e}goire Del{\'e}tang, Tim Genewein, Jordi Grau-Moya,
  R{\'o}bert Csord{\'a}s, Mehdi Bennani, Shane Legg, and Joel Veness.
\newblock Randomized positional encodings boost length generalization of
  transformers.
\newblock In Anna Rogers, Jordan Boyd-Graber, and Naoaki Okazaki (eds.),
  \emph{Proceedings of the 61st Annual Meeting of the Association for
  Computational Linguistics (Volume 2: Short Papers)}, pp.\  1889--1903,
  Toronto, Canada, July 2023. Association for Computational Linguistics.
\newblock \doi{10.18653/v1/2023.acl-short.161}.
\newblock URL \url{https://aclanthology.org/2023.acl-short.161}.

\bibitem[Saffran et~al.(2007)Saffran, Pollak, Seibel, and Shkolnik]{saffran}
Jenny Saffran, Seth Pollak, Rebecca Seibel, and Anna Shkolnik.
\newblock Dog is a dog is a dog: Infant rule learning is not specific to
  language.
\newblock \emph{Cognition}, 105:\penalty0 669--80, 12 2007.
\newblock \doi{10.1016/j.cognition.2006.11.004}.

\bibitem[Sainz et~al.(2023)Sainz, Campos, Garc{\'\i}a-Ferrero, Etxaniz,
  de~Lacalle, and Agirre]{sainz2023nlpevaluationtroubleneed}
Oscar Sainz, Jon Campos, Iker Garc{\'\i}a-Ferrero, Julen Etxaniz, Oier~Lopez
  de~Lacalle, and Eneko Agirre.
\newblock {NLP} evaluation in trouble: On the need to measure {LLM} data
  contamination for each benchmark.
\newblock In Houda Bouamor, Juan Pino, and Kalika Bali (eds.), \emph{Findings
  of the Association for Computational Linguistics: EMNLP 2023}, pp.\
  10776--10787, Singapore, December 2023. Association for Computational
  Linguistics.
\newblock \doi{10.18653/v1/2023.findings-emnlp.722}.
\newblock URL \url{https://aclanthology.org/2023.findings-emnlp.722}.

\bibitem[Schaeffer et~al.(2023)Schaeffer, Miranda, and
  Koyejo]{schaeffer2023emergent}
Rylan Schaeffer, Brando Miranda, and Sanmi Koyejo.
\newblock Are emergent abilities of large language models a mirage?
\newblock In \emph{Thirty-seventh Conference on Neural Information Processing
  Systems}, 2023.
\newblock URL \url{https://openreview.net/forum?id=ITw9edRDlD}.

\bibitem[Schmidhuber(1990)]{Schmidhuber1990TowardsCL}
Jurgen Schmidhuber.
\newblock Towards compositional learning in dynamic networkstechnical report,
  1990.
\newblock URL \url{https://api.semanticscholar.org/CorpusID:17783975}.

\bibitem[Shanahan et~al.(2020)Shanahan, Nikiforou, Creswell, Kaplanis, Barrett,
  and Garnelo]{shanahan2020explicitly}
Murray Shanahan, Kyriacos Nikiforou, Antonia Creswell, Christos Kaplanis, David
  Barrett, and Marta Garnelo.
\newblock An explicitly relational neural network architecture.
\newblock In \emph{Proceedings of the 37th International Conference on Machine
  Learning}, ICML'20. JMLR.org, 2020.

\bibitem[Shepard(1987)]{doi:10.1126/science.3629243}
Roger~N. Shepard.
\newblock Toward a universal law of generalization for psychological science.
\newblock \emph{Science}, 237\penalty0 (4820):\penalty0 1317--1323, 1987.
\newblock \doi{10.1126/science.3629243}.
\newblock URL \url{https://www.science.org/doi/abs/10.1126/science.3629243}.

\bibitem[Shi et~al.(2021)Shi, Gao, Tian, Chen, and Zhao]{shi2022learning}
Hui Shi, Sicun Gao, Yuandong Tian, Xinyun Chen, and Jishen Zhao.
\newblock Learning bounded context-free-grammar via lstm and the transformer:
  Difference and explanations.
\newblock In \emph{AAAI Conference on Artificial Intelligence}, 2021.
\newblock URL \url{https://api.semanticscholar.org/CorpusID:245329622}.

\bibitem[Siegelmann \& Sontag(1995)Siegelmann and Sontag]{SIEGELMANN1995132}
H.T. Siegelmann and E.D. Sontag.
\newblock On the computational power of neural nets.
\newblock \emph{Journal of Computer and System Sciences}, 50\penalty0
  (1):\penalty0 132--150, 1995.
\newblock ISSN 0022-0000.
\newblock \doi{https://doi.org/10.1006/jcss.1995.1013}.
\newblock URL
  \url{https://www.sciencedirect.com/science/article/pii/S0022000085710136}.

\bibitem[Sikarwar et~al.(2022)Sikarwar, Patel, and
  Goyal]{sikarwar2022transformers}
Ankur Sikarwar, Arkil Patel, and Navin Goyal.
\newblock When can transformers ground and compose: Insights from compositional
  generalization benchmarks.
\newblock In Yoav Goldberg, Zornitsa Kozareva, and Yue Zhang (eds.),
  \emph{Proceedings of the 2022 Conference on Empirical Methods in Natural
  Language Processing}, pp.\  648--669, Abu Dhabi, United Arab Emirates,
  December 2022. Association for Computational Linguistics.
\newblock \doi{10.18653/v1/2022.emnlp-main.41}.
\newblock URL \url{https://aclanthology.org/2022.emnlp-main.41}.

\bibitem[Singh et~al.(2023)Singh, Zhang, Wang, Wang, Xiong, Du, and
  Chen]{singh2023coarsetofine}
Harman Singh, Pengchuan Zhang, Qifan Wang, Mengjiao Wang, Wenhan Xiong, Jingfei
  Du, and Yu~Chen.
\newblock Coarse-to-fine contrastive learning in image-text-graph space for
  improved vision-language compositionality.
\newblock In Houda Bouamor, Juan Pino, and Kalika Bali (eds.),
  \emph{Proceedings of the 2023 Conference on Empirical Methods in Natural
  Language Processing}, pp.\  869--893, Singapore, December 2023. Association
  for Computational Linguistics.
\newblock \doi{10.18653/v1/2023.emnlp-main.56}.
\newblock URL \url{https://aclanthology.org/2023.emnlp-main.56}.

\bibitem[Sinha et~al.(2019)Sinha, Sodhani, Dong, Pineau, and
  Hamilton]{sinha2019clutrr}
Koustuv Sinha, Shagun Sodhani, Jin Dong, Joelle Pineau, and William~L.
  Hamilton.
\newblock {CLUTRR}: A diagnostic benchmark for inductive reasoning from text.
\newblock In Kentaro Inui, Jing Jiang, Vincent Ng, and Xiaojun Wan (eds.),
  \emph{Proceedings of the 2019 Conference on Empirical Methods in Natural
  Language Processing and the 9th International Joint Conference on Natural
  Language Processing (EMNLP-IJCNLP)}, pp.\  4506--4515, Hong Kong, China,
  November 2019. Association for Computational Linguistics.
\newblock \doi{10.18653/v1/D19-1458}.
\newblock URL \url{https://aclanthology.org/D19-1458}.

\bibitem[Smolensky \& Legendre(2006)Smolensky and Legendre]{cogscismolensky}
Paul Smolensky and G\'{e}raldine Legendre.
\newblock \emph{The Harmonic Mind: From Neural Computation to
  Optimality-Theoretic GrammarVolume I: Cognitive Architecture (Bradford
  Books)}.
\newblock The MIT Press, 2006.
\newblock ISBN 0262195267.

\bibitem[Smolensky et~al.(2022{\natexlab{a}})Smolensky, McCoy, Fernandez,
  Goldrick, and Gao]{Smolensky2022NeurocompositionalCI}
Paul Smolensky, R.~Thomas McCoy, Roland Fernandez, Matthew Goldrick, and
  Jianfeng Gao.
\newblock Neurocompositional computing in human and machine intelligence: A
  tutorial.
\newblock Technical Report MSR-TR-2022-5, Microsoft, May 2022{\natexlab{a}}.
\newblock URL
  \url{https://www.microsoft.com/en-us/research/publication/neurocompositional-computing-in-human-and-machine-intelligence-a-tutorial/}.
\newblock 52 pages main text, 78 pages total, 11 figures, 2 Appendices, 239
  references. For a short presentation of some of this material, see
  https://arxiv.org/abs/2205.01128 (to appear in AI Magazine).

\bibitem[Smolensky et~al.(2022{\natexlab{b}})Smolensky, McCoy, Fernandez,
  Goldrick, and Gao]{smolensky2022neurocompositional}
Paul Smolensky, Richard McCoy, Roland Fernandez, Matthew Goldrick, and Jianfeng
  Gao.
\newblock Neurocompositional computing: From the central paradox of cognition
  to a new generation of ai systems.
\newblock \emph{AI Magazine}, 43\penalty0 (3):\penalty0 308--322, Sep.
  2022{\natexlab{b}}.
\newblock \doi{10.1002/aaai.12065}.
\newblock URL
  \url{https://ojs.aaai.org/aimagazine/index.php/aimagazine/article/view/18599}.

\bibitem[Spilsbury et~al.(2024)Spilsbury, Marttinen, and
  Ilin]{spilsbury2023improved}
Sam Spilsbury, Pekka Marttinen, and Alexander Ilin.
\newblock Generating demonstrations for in-context compositional generalization
  in grounded language learning.
\newblock In \emph{Proceedings of the 2024 Conference on Empirical Methods in
  Natural Language Processing}, Miami, FL, USA, November 2024. Association for
  Computational Linguistics.

\bibitem[Subramanian et~al.(2020)Subramanian, Bogin, Gupta, Wolfson, Singh,
  Berant, and Gardner]{https://doi.org/10.48550/arxiv.2005.00724}
Sanjay Subramanian, Ben Bogin, Nitish Gupta, Tomer Wolfson, Sameer Singh,
  Jonathan Berant, and Matt Gardner.
\newblock Obtaining faithful interpretations from compositional neural
  networks, 2020.
\newblock URL \url{https://arxiv.org/abs/2005.00724}.

\bibitem[Szabó(2022)]{szabo2004compositionality}
Zoltán~Gendler Szabó.
\newblock {Compositionality}.
\newblock In Edward~N. Zalta and Uri Nodelman (eds.), \emph{The {Stanford}
  Encyclopedia of Philosophy}. Metaphysics Research Lab, Stanford University,
  {F}all 2022 edition, 2022.

\bibitem[Thagard(2023)]{sep-cognitive-science}
Paul Thagard.
\newblock {Cognitive Science}.
\newblock In Edward~N. Zalta and Uri Nodelman (eds.), \emph{The {Stanford}
  Encyclopedia of Philosophy}. Metaphysics Research Lab, Stanford University,
  {W}inter 2023 edition, 2023.

\bibitem[Tokmakov et~al.(2019)Tokmakov, Wang, and Hebert]{9010671}
Pavel Tokmakov, Yu-Xiong Wang, and Martial Hebert.
\newblock Learning compositional representations for few-shot recognition.
\newblock In \emph{2019 IEEE/CVF International Conference on Computer Vision
  (ICCV)}, pp.\  6371--6380, 2019.
\newblock \doi{10.1109/ICCV.2019.00647}.

\bibitem[Valkov et~al.(2018)Valkov, Chaudhari, Srivastava, Sutton, and
  Chaudhuri]{10.5555/3327546.3327547}
Lazar Valkov, Dipak Chaudhari, Akash Srivastava, Charles Sutton, and Swarat
  Chaudhuri.
\newblock Houdini: lifelong learning as program synthesis.
\newblock In \emph{Proceedings of the 32nd International Conference on Neural
  Information Processing Systems}, NIPS'18, pp.\  8701–8712, Red Hook, NY,
  USA, 2018. Curran Associates Inc.

\bibitem[Valvoda et~al.(2022)Valvoda, Saphra, Rawski, Williams, and
  Cotterell]{valvoda-etal-2022-benchmarking}
Josef Valvoda, Naomi Saphra, Jonathan Rawski, Adina Williams, and Ryan
  Cotterell.
\newblock Benchmarking compositionality with formal languages.
\newblock In Nicoletta Calzolari, Chu-Ren Huang, Hansaem Kim, James
  Pustejovsky, Leo Wanner, Key-Sun Choi, Pum-Mo Ryu, Hsin-Hsi Chen, Lucia
  Donatelli, Heng Ji, Sadao Kurohashi, Patrizia Paggio, Nianwen Xue, Seokhwan
  Kim, Younggyun Hahm, Zhong He, Tony~Kyungil Lee, Enrico Santus, Francis Bond,
  and Seung-Hoon Na (eds.), \emph{Proceedings of the 29th International
  Conference on Computational Linguistics}, pp.\  6007--6018, Gyeongju,
  Republic of Korea, October 2022. International Committee on Computational
  Linguistics.
\newblock URL \url{https://aclanthology.org/2022.coling-1.525}.

\bibitem[Wang et~al.(2024)Wang, Zhang, Su, and Zhu]{10444954}
Liyuan Wang, Xingxing Zhang, Hang Su, and Jun Zhu.
\newblock A comprehensive survey of continual learning: Theory, method and
  application.
\newblock \emph{IEEE Transactions on Pattern Analysis and Machine
  Intelligence}, 46\penalty0 (8):\penalty0 5362--5383, 2024.
\newblock \doi{10.1109/TPAMI.2024.3367329}.

\bibitem[Wei et~al.(2022)Wei, Tay, Bommasani, Raffel, Zoph, Borgeaud, Yogatama,
  Bosma, Zhou, Metzler, Chi, Hashimoto, Vinyals, Liang, Dean, and
  Fedus]{wei2022emergentabilitieslargelanguage}
Jason Wei, Yi~Tay, Rishi Bommasani, Colin Raffel, Barret Zoph, Sebastian
  Borgeaud, Dani Yogatama, Maarten Bosma, Denny Zhou, Donald Metzler, Ed~H.
  Chi, Tatsunori Hashimoto, Oriol Vinyals, Percy Liang, Jeff Dean, and William
  Fedus.
\newblock Emergent abilities of large language models.
\newblock \emph{Transactions on Machine Learning Research}, 2022.
\newblock ISSN 2835-8856.
\newblock URL \url{https://openreview.net/forum?id=yzkSU5zdwD}.
\newblock Survey Certification.

\bibitem[Wei et~al.(2024)Wei, Wang, Schuurmans, Bosma, Ichter, Xia, Chi, Le,
  and Zhou]{wei2023chainofthought}
Jason Wei, Xuezhi Wang, Dale Schuurmans, Maarten Bosma, Brian Ichter, Fei Xia,
  Ed~H. Chi, Quoc~V. Le, and Denny Zhou.
\newblock Chain-of-thought prompting elicits reasoning in large language
  models.
\newblock In \emph{Proceedings of the 36th International Conference on Neural
  Information Processing Systems}, NIPS '22, Red Hook, NY, USA, 2024. Curran
  Associates Inc.
\newblock ISBN 9781713871088.

\bibitem[Wu et~al.(2021)Wu, Kreiss, Ong, and Potts]{wu2021reascan}
Zhengxuan Wu, Elisa Kreiss, Desmond~C. Ong, and Christopher Potts.
\newblock Rea{SCAN}: Compositional reasoning in language grounding.
\newblock \emph{NeurIPS 2021 Datasets and Benchmarks Track}, 2021.
\newblock URL \url{https://openreview.net/forum?id=Rtquf4Jk0jN}.

\bibitem[Wu et~al.(2023)Wu, Manning, and Potts]{wu2023recogs}
Zhengxuan Wu, Christopher~D. Manning, and Christopher Potts.
\newblock {R}e{COGS}: How incidental details of a logical form overshadow an
  evaluation of semantic interpretation.
\newblock \emph{Transactions of the Association for Computational Linguistics},
  11:\penalty0 1719--1733, 2023.
\newblock \doi{10.1162/tacl_a_00623}.
\newblock URL \url{https://aclanthology.org/2023.tacl-1.96}.

\bibitem[Xu et~al.(2023{\natexlab{a}})Xu, Chai, and
  Kordjamshidi]{xu2023gipcolgraphinjectedsoftprompting}
Guangyue Xu, Joyce Chai, and Parisa Kordjamshidi.
\newblock {GIPCOL}: Graph-injected soft prompting for compositional zero-shot
  learning.
\newblock \emph{2024 IEEE/CVF Winter Conference on Applications of Computer
  Vision (WACV)}, pp.\  5762--5771, 2023{\natexlab{a}}.
\newblock URL \url{https://api.semanticscholar.org/CorpusID:265128547}.

\bibitem[Xu et~al.(2023{\natexlab{b}})Xu, Kordjamshidi, and
  Chai]{xu2023metarevisionmetalearningretrievalvisually}
Guangyue Xu, Parisa Kordjamshidi, and Joyce Chai.
\newblock {M}eta{R}e{V}ision: Meta-learning with retrieval for visually
  grounded compositional concept acquisition.
\newblock In Houda Bouamor, Juan Pino, and Kalika Bali (eds.), \emph{Findings
  of the Association for Computational Linguistics: EMNLP 2023}, pp.\
  12224--12236, Singapore, December 2023{\natexlab{b}}. Association for
  Computational Linguistics.
\newblock \doi{10.18653/v1/2023.findings-emnlp.818}.
\newblock URL \url{https://aclanthology.org/2023.findings-emnlp.818}.

\bibitem[{Yelp}(2014)]{yelp2014dataset}
{Yelp}.
\newblock Yelp dataset, 2014.
\newblock URL \url{https://www.yelp.com/dataset/}.

\bibitem[Yu \& Grauman(2014)Yu and Grauman]{6909426}
Aron Yu and Kristen Grauman.
\newblock Fine-grained visual comparisons with local learning.
\newblock In \emph{2014 IEEE Conference on Computer Vision and Pattern
  Recognition}, pp.\  192--199, 2014.
\newblock \doi{10.1109/CVPR.2014.32}.

\bibitem[Yu \& Grauman(2017)Yu and Grauman]{Yu_2017}
Aron Yu and Kristen Grauman.
\newblock Semantic jitter: Dense supervision for visual comparisons via
  synthetic images.
\newblock In \emph{2017 IEEE International Conference on Computer Vision
  (ICCV)}. IEEE, October 2017.
\newblock \doi{10.1109/iccv.2017.594}.
\newblock URL \url{http://dx.doi.org/10.1109/ICCV.2017.594}.

\bibitem[Yu et~al.(2024)Yu, Kaur, Gupta, Brown-Cohen, Goyal, and
  Arora]{yu2023skillmixflexibleexpandablefamily}
Dingli Yu, Simran Kaur, Arushi Gupta, Jonah Brown-Cohen, Anirudh Goyal, and
  Sanjeev Arora.
\newblock {SKILL}-{MIX}: a flexible and expandable family of evaluations for
  {AI} models.
\newblock In \emph{The Twelfth International Conference on Learning
  Representations}, 2024.
\newblock URL \url{https://openreview.net/forum?id=Jf5gplvglq}.

\bibitem[Zheng \& Lapata(2021)Zheng and
  Lapata]{zheng-lapata-2021-compositional-generalization}
Hao Zheng and Mirella Lapata.
\newblock Compositional generalization via semantic tagging.
\newblock In Marie-Francine Moens, Xuanjing Huang, Lucia Specia, and Scott
  Wen-tau Yih (eds.), \emph{Findings of the Association for Computational
  Linguistics: EMNLP 2021}, pp.\  1022--1032, Punta Cana, Dominican Republic,
  November 2021. Association for Computational Linguistics.
\newblock \doi{10.18653/v1/2021.findings-emnlp.88}.
\newblock URL \url{https://aclanthology.org/2021.findings-emnlp.88}.

\bibitem[Zheng \& Lapata(2022)Zheng and Lapata]{zheng2022disentangled}
Hao Zheng and Mirella Lapata.
\newblock Disentangled sequence to sequence learning for compositional
  generalization.
\newblock In Smaranda Muresan, Preslav Nakov, and Aline Villavicencio (eds.),
  \emph{Proceedings of the 60th Annual Meeting of the Association for
  Computational Linguistics (Volume 1: Long Papers)}, pp.\  4256--4268, Dublin,
  Ireland, May 2022. Association for Computational Linguistics.
\newblock \doi{10.18653/v1/2022.acl-long.293}.
\newblock URL \url{https://aclanthology.org/2022.acl-long.293}.

\bibitem[Zhong et~al.(2024)Zhong, Li, Wang, Song, Wei, Lian, and
  Mao]{zhong2024benchmarkingimprovingcompositionalgeneralization}
Tianqi Zhong, Zhaoyi Li, Quan Wang, Linqi Song, Ying Wei, Defu Lian, and
  Zhendong Mao.
\newblock Benchmarking and improving compositional generalization of
  multi-aspect controllable text generation.
\newblock In Lun-Wei Ku, Andre Martins, and Vivek Srikumar (eds.),
  \emph{Proceedings of the 62nd Annual Meeting of the Association for
  Computational Linguistics (Volume 1: Long Papers)}, pp.\  6486--6517,
  Bangkok, Thailand, August 2024. Association for Computational Linguistics.
\newblock \doi{10.18653/v1/2024.acl-long.351}.
\newblock URL \url{https://aclanthology.org/2024.acl-long.351}.

\end{thebibliography}
\end{document}